\documentclass[journal]{IEEEtai}

\usepackage[colorlinks,urlcolor=blue,linkcolor=blue,citecolor=blue]{hyperref}

\usepackage{color,array}
\usepackage{graphicx}
\usepackage{amssymb}
\usepackage{amsmath}
\usepackage{relsize}
\usepackage{stmaryrd}
\usepackage{multirow}
\usepackage{makecell}
\let\labelindent\relax
\usepackage{enumitem}
\usepackage{bm}
\usepackage{tikz}
\usepackage{tikz-qtree}
\usepackage{tikz-qtree-compat}
\usepackage{tabularx}
\usepackage{svg}
\usepackage{setspace}

\usepackage{bbding}
\usepackage{pifont}
\usepackage[nointegrals]{wasysym}
\newcommand{\cmark}{\ding{51}}%
\usepackage{url}
\usepackage{booktabs}
\usepackage{rotating}

\usepackage{subcaption}
\usepackage[numbers]{natbib}
\usetikzlibrary{shapes,decorations,arrows,calc,arrows.meta,fit,positioning}
\tikzset{
	-Latex,auto,node distance =1 cm and 1 cm,semithick,
	state/.style ={ellipse, draw, minimum width = 0.7 cm},
	point/.style = {circle, draw, inner sep=0.04cm,fill,node contents={}},
	bidirected/.style={Latex-Latex,dashed},
	el/.style = {inner sep=2pt, align=left, sloped}
}

\def\independenT#1#2{\mathrel{\rlap{$#1#2$}\mkern2mu{#1#2}}}
\newcommand\independent{\protect\mathpalette{\protect\independenT}{\perp}}

\newtheorem{definition}{Definition}
\begin{document}

\title{Evaluation Methods and Measures for Causal Learning Algorithms} 
\author{Lu Cheng, Ruocheng Guo, Raha Moraffah, Paras Sheth, K. Sel\c{c}uk Candan, and Huan Liu \IEEEmembership{Fellow, IEEE}
\thanks{Ruocheng Guo is with School of Data Science, City University of Hong Kong, China. All other authors are with School of Computing, Informatics, and Decision Systems Engineering, Arizona State University, Tempe, AZ, USA. (e-mail: \{lcheng35,rmoraffa,psheth5,candan,huanliu\}@asu.edu, ruocheng.guo@cityu.edu.hk)}}
\markboth{Journal of IEEE Transactions on Artificial Intelligence, Vol. 00, No. 0, Month 2022}
{Lu Cheng \MakeLowercase{\textit{et al.}}: Evaluation Methods and Measures for Causal Learning Algorithms}

\maketitle

\begin{abstract}
The convenient access to copious multi-faceted data has encouraged machine learning researchers to reconsider correlation-based learning and embrace the opportunity of causality-based learning, i.e., causal machine learning (causal learning). Recent years have therefore witnessed great effort in developing causal learning algorithms aiming to help AI achieve human-level intelligence. Due to the lack-of ground-truth data, one of the biggest challenges in current causal learning research is algorithm evaluations. This largely impedes the cross-pollination of AI and causal inference, and hinders the two fields to benefit from the advances of the other. To bridge from conventional causal inference (i.e., based on statistical methods) to causal learning with big data (i.e., the intersection of causal inference and machine learning), in this survey, we review commonly-used datasets, evaluation methods, and measures for causal learning using an evaluation pipeline similar to conventional machine learning. We focus on the two fundamental causal-inference tasks and causality-aware machine learning tasks. Limitations of current evaluation procedures are also discussed. We then examine popular causal inference tools/packages and conclude with primary challenges and opportunities for benchmarking causal learning algorithms in the era of big data. The survey seeks to bring to the forefront the urgency of developing publicly available benchmarks and consensus-building standards for causal learning evaluation with observational data. In doing so, we hope to broaden the discussions and facilitate collaboration to advance the innovation and application of causal learning.
\end{abstract}

\begin{IEEEImpStatement}
Causal learning goes beyond machine learning due to its power of uncovering data generating processes. Causality relates to crucial open problems in machine learning. On the opposite, machine learning contributes to addressing fundamental challenges in causal inference. One key challenge of causal learning is that the research domain lacks public benchmark resources to support principled evaluation of research contributions. Our goal is to promote objectivity, reproducibility, fairness, collaboration, and awareness of bias in causal learning research. Arguing that this goal can only be achieved through systematic, objective, and transparent evaluation, in this survey, we provide a comprehensive review of the evaluation of fundamental tasks in causal inference and causality-aware machine learning tasks. Similar to the evaluation in conventional machine learning, the causal evaluation pipeline includes the evaluation protocols, metrics, datasets, and popular causal tools/packages. We also seek to expedite the marriage of causality and machine learning via discussions of prominent open problems and challenges.
\end{IEEEImpStatement}
\section{Introduction}
\IEEEPARstart{M}{achine} learning (ML) is a key pillar of artificial intelligence (AI). The unmatched availability of big data unleashed the unprecedented power of ML to support situational awareness and decision making. While witnessing the exceptional success of ML technologies in various applications, users started to notice a critical shortcoming of ML. Traditional ML techniques can learn correlation-based patterns and relationships from data. {\it Unfortunately, correlation is a poor substitute for causality as in many cases, data may contain spurious correlations}~\cite{pearl2009causality,cheng2019robust}.
Causal inference with observational data -- data that have been generated by something other than randomized experiments -- offers a promising alternative to correlation-based learning. Causal inference excels at uncovering the underlying causal mechanisms, leading to its wide applications in a myriad of high-impact domains, such as education~\cite{lalonde1986evaluating,dehejia1999causal,heckerman2006bayesian,hill2011bayesian}, medical science~\cite{mani2000causal,cross2013identification}, economics~\cite{imbens2004nonparametric}, epidemiology~\cite{robins2000marginal,hernan2018causal}, meteorology~\cite{ebert2012causal}, and environmental health~\cite{li2014discovering}. Therefore, learning causality is significant for AI achieving human-level intelligence~\cite{pearl2018theoretical}.

The conventional way to understand causality is to use interventions and/or randomized controlled trials (RCTs)~\cite{rubin2005causal,guo2020survey}. In many situations, however, these are time-consuming, impractical, or sometimes even unethical~\cite{kallus2018confounding,guo2020survey,cheng2019practical}. Attention has been drawn to the recent availability of big observational data in all walks of life as they provide new opportunities for learning causality, without the disadvantages of RCTs. While being relatively recent, causal ML (causal learning, CL) with observational data is emerging as a vibrant field with new opportunities and domain specific challenges. CL ``seeks to model the effect of interventions and distribution changes with a combination of data-driven learning and assumptions not already included in the statistical description of system'' \cite{scholkopf2021toward}. Note that causal inference is relevant to methods that perform inference (effect estimation) and structural learning (causal discovery), whereas CL is more frequently mentioned along with methods that leverage ML to improve causal inference tasks or use causality to resolve some limitations of current ML methods. In this work, we focus on the two fundamental problems in causal inference: {\bf (1)} {\em learning causal effects (i.e., estimating causal effect of a treatment on an important outcome)} and {\bf (2)} {\em learning causal structure (i.e., examining whether a certain set of causal relations exists between variables)}~\cite{scholkopf2019causality}. Answering these challenging causal questions is starting to become feasible as large datasets may contain sufficient samples from the joint distribution of the observed variables~\cite{rubin2005causal,guo2019learning}. 

A primary obstacle that impedes research developments in CL is the lack-of public benchmark resources to support principled evaluation. Standardized evaluation played a major role in the early days of ML research. Successful early benchmarking efforts, such as UCI ML\footnote{http://archive.ics.uci.edu/ml/} and UCI KDD\footnote{https://kdd.ics.uci.edu/} repositories, not only helped guide the development of efficient and effective ML algorithms, but also encouraged collaborative research and paved the way for the recent breakthroughs in deep learning. Unfortunately, given the different learning objectives of conventional ML and CL, these existing benchmarks are not applicable to CL. 

The overarching goal is, therefore, to enable the advancement of CL research, and promote objectivity, reproducibility, fairness, collaboration, and awareness of bias in CL research. We argue that this goal can only be achieved through systematic, objective, and transparent evaluation of CL models and algorithms. In this survey, we aim to provide a comprehensive review for the evaluations of the two fundamental tasks in causal inference and causality-aware ML tasks, such as causal interpretability. We summarize the results in Table \ref{ontology}. Similar to evaluation in conventional ML, the evaluation pipeline for CL includes the evaluation protocols, metrics, datasets, and popular benchmarking tools/packages. Under each task, we then discuss the limitations of current evaluation procedures. This survey also aims to help set agenda in future research on benchmarking CL algorithms by examining prominent open problems and challenges in current evaluation pipeline for CL.

\noindent\textbf{Difference from Previous Work.}
Here, we first briefly summarize existing surveys on causal inference and CL and then highlight their differences from this work.

Guo et al.~\cite{guo2020survey} focus on reviewing the methodologies for causal effect estimation and causal structure learning, as well as discussing the connections between causal inference and ML.
Yao et al.~\cite{yao2020survey} specifically survey causal effect estimation methods and tools under the potential outcome framework.
Another survey by Spirtes and Zhang~\cite{spirtes2016causal} reviews semi-parametric score-based methods for learning causal structure with i.i.d. (independent and identically distributed) and time-series observational data.
To bridge from ML to Artificial General Intelligence, Sch{\"o}lkopf~\cite{scholkopf2019causality} shows that many challenging problems in ML and AI are inherently related to causality. The author especially examines where the links between AI and graphical causal inference have been and should be established.
The rapid growth of time-series data and the unique challenges it brings in causal studies have led to surveys such as \cite{moraffah2021causal} reviewing problems, methods, and evaluation related to causal time series analysis. 
In causal ML, Chen et al.~\cite{chen2020bias} summarize different types of bias in recommendation systems and review methods that aim to mitigate such bias by leveraging causal inference theories. Moraffah et al.~\cite{moraffah2020causal} survey the methods that leverage causality to enable the interpretability of ML models. All previously mentioned surveys center on reviewing the theories and methods in CL without an in-depth discussion about the evaluation.

The most related work is by Shimoni et al. \cite{shimoni2018benchmarking} and  Mooiji et al. \cite{shimoni2018benchmarking}. The former introduces a benchmarking framework for causal effect estimation. It summarizes evaluation metrics and data generation with ground-truth effect. The second work surveys methods and benchmarks for causal structure learning. Compared to prior studies, this survey aims to present a comprehensive review of the evaluation protocols, \textit{evaluation metrics}, datasets, and causal tools/packages that have been widely used to benchmark causal effect estimation and causal structure learning. In addition, with the growing interest in causality in AI community, we also examine existing framework for evaluating causality-aware ML tasks via several representative examples such as causal interpretability and fairness. Therefore, this survey complements existing surveys 1) focused on reviewing causal theories and methodologies; and 2) benchmarking one fundamental task in causal inference, i.e., either causal structure learning or causal effect estimation. 

\noindent\textbf{Intended Audience and Paper Organization.} This survey will most benefit researchers and practitioners who have basic knowledge of causal inference and would like to develop or apply CL algorithms, but often find the evaluation rather challenging. It will also be useful whoever is interested in knowing the differences between evaluating standard ML and CL algorithms. The rest of the survey is organized as follows: We discuss evaluation frameworks for causal effect estimation in Section 2, causal structure learning in Section 3, and causal ML tasks in Section 4. We then summarize and compare existing tools and packages for causal inference in Section 5. Section 6 delineates prominent open problems and challenges and the last section concludes the survey.

\begin{table*}[]
\centering
\resizebox{\textwidth}{!}{%
\begin{tabular}{|p{2mm}|c|c|c|c|l|c|}
\hline
\textbf{} &
  \multicolumn{2}{c|}{\textbf{\begin{tabular}[c]{@{}c@{}}Causal Effect Estimation\end{tabular}}} &
  \textbf{Causal Structure Learning} &
  \multicolumn{2}{c|}{\textbf{\begin{tabular}[c]{@{}c@{}}Causal Interpretability\\ and Fairness\end{tabular}}} &
  \textbf{\begin{tabular}[c]{@{}c@{}}Unbiased Interactive\\ ML\end{tabular}} \\ \hline
\multirow{3}{*}{\textbf{\begin{turn}{90}Metrics\end{turn}}} &
  \textbf{\begin{tabular}[c]{@{}c@{}}Standard \\ Effect \\ Metrics\end{tabular}} &
  \begin{tabular}[c]{@{}c@{}}MAE, MSE, \\ RMSE, PEHE, \\ Policy Risk\end{tabular} &
  \multirow{3}{*}{\begin{tabular}[c]{@{}c@{}}SHD, SID, \\ Frobenius Norm,\\ Precision, Recall, F1, \\ TPR, FPR, MSE, AUC,\\ Precision-Recall Curve,\\ FPR-TPR Curve,\\
  TVD, KL-Divergence,\\
  F-test\end{tabular}} &
  \textbf{\begin{tabular}[c]{@{}c@{}}Counterfactual \\ Explanation\end{tabular}} &
  \multicolumn{1}{c|}{\begin{tabular}[c]{@{}c@{}}Sparsity, \\ Interpretability,\\ Speed, Proximity, \\ Diversity, \\ Visual Linguistic\end{tabular}} &
  \begin{tabular}[c]{@{}c@{}}NDCG@K, MAP@K, \\ARP@K, APLT@K\end{tabular}\\ \cline{2-3} \cline{5-6}
 &
  \textbf{\begin{tabular}[c]{@{}c@{}}Heterogenous \\ Effect \\ Metrics\end{tabular}} &
  \begin{tabular}[c]{@{}c@{}}$Uplift_{Coef}$, \\ $Qini_{Coef}$\end{tabular} &
  &
  \multirow{2}{*}{\textbf{Fairness}} &
  \multirow{2}{*}{\begin{tabular}[c]{@{}c@{}}FACE, FACT,\\Counterfactual Fairness,\\PC-Fairness,\\Ctf-DE, Ctf-IE, Ctf-SE\end{tabular}} & \\ \cline{2-3} 
 &
  \textbf{\begin{tabular}[c]{@{}c@{}}Time Series \\ Metrics\end{tabular}} &
  \begin{tabular}[c]{@{}c@{}}Standard and Heterogeneous \\ Effect Metrics, F-Test, T-Test\end{tabular} &
  &
  &
   &
  \\ \hline
\multirow{2}{*}{\begin{turn}{90}\textbf{Procedures}\end{turn}} &
  \textbf{\begin{tabular}[c]{@{}c@{}}With \\ Ground \\ Truth\end{tabular}} &
  \multicolumn{1}{l|}{\begin{tabular}[c]{@{}l@{}}Observational data with \\known effect; observational \\and experimental data pairs;\\ sampling from observational\\ data; sampling from synthetic\\ data;
  sampling from RCTs\end{tabular}} &
  \multicolumn{1}{l|}{\multirow{2}{*}{\begin{tabular}[c]{@{}l@{}}A transductive \\ setting where we have the \\ ground-truth causal\\ graph and estimated graph\end{tabular}}} &
  \textbf{Transductive} &
  \begin{tabular}[c]{@{}l@{}}Training on a regular dataset \\ and testing on generated \\ counterfactuals\end{tabular} &
  \multicolumn{1}{l|}{\multirow{2}{*}{\begin{tabular}[c]{@{}l@{}}Training set comes \\from a biased source\\ whereas test set comes\\ from an unbiased source\end{tabular}}} \\ \cline{2-3} \cline{5-6}
 &
  \textbf{\begin{tabular}[c]{@{}c@{}}Without \\ Ground \\ Truth\end{tabular}} &
  \multicolumn{1}{l|}{\begin{tabular}[c]{@{}l@{}}Evaluation is possible if\\ subset of the data is from RCTs\end{tabular}} &
  \multicolumn{1}{l|}{} &
  \textbf{Inductive} &
  \begin{tabular}[c]{@{}l@{}}Generating counterfactual\\  explanations for an unseen \\ instance\end{tabular} &
  \multicolumn{1}{l|}{} \\ \hline
\begin{turn}{90}
  \textbf{Dataset}
\end{turn} &
  \multicolumn{2}{c|}{\begin{tabular}[c]{@{}c@{}}Under Unconfoundedness Assumption,\\ Natural Experiments, RCTs\end{tabular}} &
  \begin{tabular}[c]{@{}c@{}}Causal Direction, \\ Causal Graphs,\\ Time Series Datasets\end{tabular} &
  \multicolumn{2}{c|}{Image, Text, Tabular} &
  \begin{tabular}[c]{@{}c@{}}Semi-Synthetic \\datasets, RCTs\end{tabular}  \\ \hline
\end{tabular}
}
\caption{Summary of metrics, procedures, datasets for evaluating CL approaches.}
\label{ontology}
\end{table*}
\section{Benchmarking Causal Effects Estimation}
We specify the pipelines for evaluating the first fundamental task in causal inference -- causal effect estimation. We begin by defining the problem in the Neyman/Rubin Potential Outcome Framework \cite{imbens2015causal} and then introduce the benchmarking datasets, evaluation procedures, and metrics.

Many empirical analyses are motivated by the necessity to estimate the causal effects of a binary treatment on the outcome of interest. Let $\bm{x}_i$, $t_i\in\{0,1\}$, $y_i$ be the features (i.e., background variables), treatment assignment with $t_i=1$ being under treated and $t_i=0$ under control, and outcome of unit $i$. Estimating causal effects is defined as 
\begin{definition}[Causal Effects Estimation]
	\textit{Given $n$ units $\{(\bm{x}_{1},t_{1},y_1),...,(\bm{x}_{n},t_{n},y_{n})\}$, estimating causal effects is to quantify the changes of $Y$ as we alter the treatment assignment from $0$ to $1$.}
\end{definition}
\noindent Under different contexts, effects can be estimated within the entire population, a subpopulation that is defined by background variables, an unknown subpopulation, or an individual. Average Treatment Effect (ATE) $\tau$ is typically used for assessing a population represented by the distribution of $X$:
\begin{equation}
\tau = \mathbb{E}_X[\tau(\bm{x})]=\mathbb{E}_X[Y|do(T=1)]-\mathbb{E}_X[Y|do(T=0)], 
\end{equation}
where $do(T=1)$ and $do(T=0)$ indicate that the treatment assignment is ``treated'' and ``control'', respectively. When the population is heterogeneous, ATE can be misleading as same treatment may affect individuals differently. Under heterogeneity, a common assumption requires that each subpopulation is defined by a set of features, i.e., Conditional ATE (CATE):
\begin{equation}
\begin{aligned}
CATE: \tau(\bm{x}) = \mathbb{E}[Y|do(t=1),\bm{x}] - \mathbb{E}[Y|do(t=0),\bm{x}].
\label{eq:CATE}
\end{aligned}
\end{equation}
An Individual Treatment Effect (ITE) is a contrast between potential outcomes of a unit:
\begin{equation}
    \tau_i=Y_i(do(t=1))-Y_i(do(t=0)).
\end{equation}
Note that $\tau_i$ is not necessarily equal to $\tau(\bm{x})$ as the latter is an average over a subpopulation. The goal of the {\em causal effects estimation} task is to learn a function $\hat{\tau}$ that estimates ATE or CATE, depending on the degree of homogeneity of the population, for binary treatment options: $T=0$ and $T=1$.
\subsection{Evaluation Metrics}
Evaluation metrics for causal effect estimation can be categorized into metrics for standard and heterogeneous effect estimations, as well as for time-series effect estimation. We review popular metrics in each category.
\subsubsection{Metrics for Standard Causal Effect Estimation} As effect is typically continuous, most of the metrics are directly adapted from those for regression in ML, e.g., Root Mean Square Error (RMSE), Mean Absolute Error (MAE), precision in estimation of heterogeneous effect (PEHE), and Policy Risk.
Given the ground-truth ATE $\tau$ and the predicted ATE $\hat{\tau}$, MAE of estimating ATE is defined as
\begin{equation}
	\epsilon_{MAE\_ATE} = \frac{1}{M}\sum_{j=1}^M|\tau_j-\hat{\tau}_j|,
\end{equation}
where $M$ stands for the number of experiments and $j$ index of the experiment. There are other similar metrics such as \textit{mean squared error} (MSE) and RMSE~\cite{hartford2017deep,puli2019generalized}:
\begin{equation}
\begin{aligned}
    	\epsilon_{MSE\_ATE}  &= \frac{1}{M}\sum_{j=1}^M(\tau_j-\hat{\tau}_j)^2,\\
		\epsilon_{RMSE\_ATE}  &= \sqrt{\frac{1}{M}\sum_{j=1}^M(\tau_j-\hat{\tau}_j)^2}.
\end{aligned}
\end{equation}

\noindent\textit{PEHE} is used for estimating CATE and is defined as
\begin{equation}
\epsilon_{PEHE} = \frac{1}{n}\sum_{i=1}^n(y_i^1-y_i^0-\hat{\tau}(\bm{x}_i))^2,
\end{equation}
where $y_i^1$ and $y_i^0$ are the two potential outcomes of the $i$-th unit, $y_i^1-y_i^0$ denotes the ground-truth CATE, and $\hat{\tau}(\cdot)$ is the learned function. When the true ITE is unknown but a subset of the dataset comes from an RCT, one can use \noindent\textit{Policy Risk}, defined as the average loss in value when treatment is assigned based on the ITE estimator-guided policy \cite{shalit2017estimating,rakesh2018linked}:
\begin{equation}
\begin{aligned}
    \text{Policy Risk} = 1 &- \big(\mathbb{E}[\tilde{y}^1_i|t_i=1,\pi_i=1]p(\pi_i=1) \\ &+\mathbb{E}[\tilde{y}_i^0|t_i=0,\pi_i=0]p(\pi_i=0)\big),
    \label{PR}
\end{aligned}
\end{equation}
where $\pi_i=1$ denotes the policy to treat, and to not treat $\pi_i=0$ otherwise. $\tilde{y}_i$ is the factual outcome scaled between $[0,1]$. The second component in Eq. \ref{PR} represents the expected potential outcome, i.e., the weighted sum of expectations of two potential outcomes.
\subsubsection{Metrics for Heterogeneous Effect Estimation} Evaluating heterogeneous effects with binary treatment typically follows the uplift modeling literature~\cite{DBLP:conf/sdm/ZhaoFS17,gerardycausal}. Uplift modeling is an ML approach that employs the Potential Outcome framework to estimate ITE in order to customize treatment assignments for different units. For example, companies use uplift models to differ between customers who buy product because of the campaign and those who will buy anyway. The evaluation metrics are defined over a curve measuring the performance of algorithms: The $x$-axis of the curve represents the number of units or the percentile of the population sorted by each estimated ITE (the larger the estimated ITE, the smaller the percentile); the $y$-axis represents gain when the treatment is assigned to the top $a$-percentile. Given population sorted by the inferred ITEs, \textit{uplift curve} (a.k.a. cumulative gain chart) measures the average cumulative gain of receiving the treatment in the first $b$ units:
\begin{equation}
    Uplift(b) = \left(\frac{Y_b^1}{N_b^1}-\frac{Y_b^0}{N_b^0}\right)(N_b^1+N_b^0),
\end{equation}
where $Y_b^1$ ($Y_b^0$) and $N_b^1$ ($N_b^0$) denote the sum of the treated (control) outcomes and the number of treated (control) units among the first $b$ units, respectively. A variant of uplift curve is the \textit{Qini curve} that is defined in terms of percentiles, instead of the absolute number of units~\cite{radcliffe2007using}:
\begin{equation}
  Qini(a) = Y_a^1 - \frac{Y_a^0 N_a^1}{N_a^0},
\end{equation}
where $Y_a^1$ ($Y_a^0$) and $N_a^1$ ($N_a^0$) denote the sum of the treated (control) outcomes and the number of treated (control) units in the first $a$ percentile of population, respectively. Similar to AUC-ROC curve, we can compute the area under the Uplift/Qini curve, referred to as the uplift/Qini coefficient:
\begin{equation}
    \begin{split}
    Uplift_{Coef} & = \sum_{b=0}^{N-1} \frac{1}{2}(Uplift(b+1)+Uplift(b)); \\
        Qini_{Coef} & = \sum_{a=0}^{N-1} \frac{1}{2}(Qini(a+1)+Qini(a)),
    \end{split}
\end{equation}
where $N$ is the size of the population.
\subsubsection{Metrics for Time Series Effect Estimation}
Time series data are sequences of real-valued data ordered over time. Causal inference for time series analysis is of particular interest because many scientific questions involve causal effect estimation and causal structure learning using chronological observations \cite{moraffah2021causal}, such as those related to medicine and social science. For example, social scientists have been interested in studying the effects of minimum wages on employment, where monthly employment rate needs to be recorded before and after a change in the minimum wage is made \cite{card1993minimum}.

Evaluation metrics for standard effect estimation (such as MAE described above) can be used for time series effect estimation at each time step $s$. To evaluate over the entire time series, we need to further take an average of the results over time, e.g., $MAE=\sum_{s=1}^S MAE_s$. In addition, we can use metrics specifically designed for sequential data such as F-Test and T-Test. F-test assesses treatment effect heterogeneity by comparing the marginal variances of two potential outcomes. Let $\hat{e}_t$ and $\hat{u}_t$ be the time dependent errors for the treated and control groups, respectively. $p$ denotes the time lag, F-Test is defined as
\begin{equation}
    F=\frac{(RSS_0-RSS_1)/p}{RSS_1/(S-2p-1)},
\end{equation}
where $RSS_0=\sum_{s=1}^S\hat{e}_t$ and $RSS_1=\sum_{s=1}^S\hat{u}_t$. To test the significance of a cause, one can also use the Unpaired T-Test (UTT) to compare the sequence of the treated group with that of the control group:
\begin{equation}
    UTT=\frac{\bar{x}_1-\bar{x}_2}{\sqrt{(\frac{1}{n_1}+\frac{1}{n_2})\big(\frac{(n_1-1)s_1^2+(n_2-1)s_2^2}{(n_1-1+n_2-1)}\big)}},
\end{equation}
where $\bar{x}_1$ and $\bar{x}_2$ are the mean of two sequences; $s_1$ and $s_2$ denote the standard deviations; and $n_1$ and $n_2$ denote the cardinalities of the sequences. 
\subsection{Evaluation Procedures}
Evaluating causal inference methods is significantly more challenging than evaluating purely associational methods. The primary obstacle in the evaluation procedure using observational data is that we typically cannot know the true causal effects. Some prior work uses observational data with known treatment effect. However, this requires that the studied phenomena are so well-understood that the causal effect is obvious \cite{mooij2016distinguishing}, limiting data availability. Another strategy uses data from pairs of observational and experimental studies to create a nearly ideal scenario for causal effect estimation with observational data \cite{dixit2016perturb}. This still suffers from low data availability. The most straightforward approach is to generate observational data from synthetic causal systems where the treatment effect is either directly known or can be easily derived from the formulation \cite{louizos2017causal,kallus2018causal}. In particular, consider a data generating process that produces binary treatment $T$, potential outcome $Y(t)$, and multiple covariates $X=\{X_1, X_2,..., X_k\}$. For each unit $i$, both potential outcomes $Y_i(1)$ and $Y_i(0)$ are measured. Given the biasing covariates $X^b\subseteq X$ of a unit and the synthetic dataset, one can begin by generating the selected treatment $T^s$ using the biasing covariates $P(T^s|X^b):=f(X^b)$. If $T^s$ is the same as the unit's corresponding treatment in the synthetic data, this unit is then added to the simulated observational dataset. This evaluation procedure is \textit{sampling from synthetic data}. However, synthetic data cannot generalize well to real-world settings.

A more recent class of existing work augments an observational study with synthetic treatment assignment and outcomes generated by a synthetic function \cite{shimoni2018benchmarking}. Given the observed covariates, we first randomly select a subset of samples which is then fed into a data generating process to simulate treatment assignment and potential outcomes. Causal models are then trained on the tuples of treatment, covariates, and factual outcome. This approach -- referred to as \textit{sampling from observational data} -- can be used to evaluate both individual- and population-level effects. The newest approach creates constructed observational data by \textit{sampling from RCTs} \cite{gentzel2021and}. It has been shown that, in expectation, this approach can create datasets equivalent to those produced by randomly sampling from empirical datasets where all potential outcomes are known. This approach is only suited for estimating population-level effects. Once the ground truth is available, evaluating causal effects estimation is similar to supervised learning algorithms. Without ground truth, evaluation is still possible if the subset of data is from RCTs, e.g., policy risk for ITE. 

The evaluation of causal effect estimation with time-series and networked data typically follow the same protocol and metrics as that with i.i.d. data. The estimated ATE (ITE) of the test set will be compared with the corresponding ground truth using MAE and MSE (PEHE).
\subsection{Benchmarking Datasets}
In general, there are two types of data used for identifying and estimating causal effects~\cite{guo2020survey}: data used under the \textit{unconfoundedness} assumption~\cite{rubin2005causal} -- observed/measured variables are sufficient to capture the causal links between treatment and outcome; and data collected from natural experiments, popular alternatives to RCTs. Below, we briefly introduce exemplar benchmarking datasets in each category. For a comprehensive description, please refer to \cite{cheng2019practical}. We first introduce data used under the unconfoundedness assumption.
\subsubsection{Datasets with Binary Treatment Variables}
Most methods in the literature of learning causal effects are evaluated on datasets with binary treatment.
\begin{itemize}[leftmargin=*]
    \item \textit{Jobs.} The research studies the effect of a job training program on the real earnings of an individual. This dataset consists of RCT conducted by LaLonde~\cite{lalonde1986evaluating} (297 treated and 425 control) and the Panel Study of Income Dynamics comparison group (2,490 control)~\cite{smith2005does}. Features are multiple demographic variables such as age and race.
    \item \textit{(Infant Health Development Program) IHDP.} 
This is a dataset with simulated treatment and outcomes, initially complied by~\cite{hill2011bayesian}, seeking to evaluate the efficacy of comprehensive early intervention in reducing the development and health problems of low birth weight, premature infants.
A commonly used simulation setting is the setting ``A'' in the NPCI package\footnote{https://github.com/vdorie/npci}. This dataset comprises 747 instances (139 treated and 608 control) \and each is associated with 25 features.
\item \textit{Atlantic Causal Inference Conference Benchmark (ACICB)}~\cite{hahn2018atlantic}. It inherits the same features as those in IHDP data~\cite{duncan1994economic}. Various settings have been adopted to synthesize the treatment and outcomes.
\item \textit{Twins.}
The Twins dataset in~\cite{almond2005costs} is used to study ITE of twins' weights on their mortality in the first year of their births. Each twin-pair is represented by 46 features relating to the parents, the pregnancy, and birth.
\item \textit{BlogCatalog} is used to study causal effect estimation with networked data. It is collected from an online social network service where users can post blogs. Treatments and outcomes are synthesized based on the observed features, the social network structures, and the homophily phenomenon. This dataset has 5,196 instances, 173,468 edges, and 8,189 observed features.
\item \textit{Amazon}~\cite{rakesh2018linked} is an extension of the dataset in ~\cite{mcauley2015inferring}. The goal is to evaluate the efficacy of positive (or negative) reviews in promoting product sales. Treatment and potential outcomes are generated based on the ratings and by matching the most similar products with different treatment assignment.
\end{itemize}

\subsubsection{Datasets with Multiple Treatments/Continuous Treatment} 
There are studies in which multiple treatments are studied, e.g., \textit{Twins-Mult}~\cite{yoon2018ganite}, \textit{TCGA}~\cite{schwab2018perfect}, and \textit{News-Mult}~\cite{schwab2018perfect}. A dataset with continuous treatment can be found in~\cite{galagate2015package}. It was collected to study the causal effect of the amount of smoking on the medical expenditure.

Datasets collected from natural experiments\footnote{In most of the cases, natural experiments only allow us to identify ATEs but not ITEs or CATEs, therefore, we can use the ATE metrics such as $\epsilon_{MAE\_ATE}$, $\epsilon_{MSE\_ATE}$, and $\epsilon_{RMSE\_ATE}$ for evaluation~\cite{hartford2017deep,puli2019generalized}.} allow us to relax the stringent unconfoundedness assumption~\cite{angrist2014mastering} as it considers the presence of hidden confounders. Such datasets include variables selected under causal knowledge such as the \emph{instrumental variable(s) (IVs)} and \emph{running variable(s)}. Below, we review the datasets with IVs~\cite{angrist1996identification} and RDD (Regression Discontinuity Design)~\cite{campbell1969reforms}. We also discuss datasets for causal time series analysis and datasets with network information given the commonness of these two types of data.
\subsubsection{Datasets with Instrumental Variables} 
 IV is a powerful tool to identify causal effect when there exists hidden confounders. We first briefly describe the core idea of IV and then describe exemplar datasets.

\begin{figure}[t]
	\centering
	\begin{tikzpicture}
	\node[state] (d) at (2,0) {$T$};
	\node[state] (i) at (0,0) {$Z$};
	\node[state,dotted] (u) at (1,1) {$\mathbf{U}$};
	
	
	\node[state] (x) at (3,1) {$\mathbf{X}$};
	\node[state] (y) at (4,0) {$Y$};
	
	\path (x) edge (d);
	\path (x) edge (y);
	\path (d) edge (y);
	\path (i) edge (d);
	\path (u) edge (d);
	\path (u) edge (y);	
	
	\end{tikzpicture}
	\caption{A causal graph of a valid IV ($Z$) when there are hidden confounders ($\mathbf{U}$). $\mathbf{X}$ stands for the observed confounders and $\mathbf{U}$ is a set of hidden confounders.}
	\label{fig:IV1}
\end{figure}
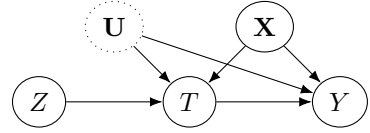

\begin{definition}[Instrumental Variable]
	\textit{Given an observed variable $Z$, observed features $\mathbf{X}$, treatment $T$ and outcome $Y$, $Z$ is a valid IV for the causal effect of $T \rightarrow Y$ iff $Z$ satisfies: (1) $Z\not\independent T|\mathbf{X}$, and (2) $Z \independent Y|\mathbf{X}, do(T)$~\cite{angrist1996identification}.}
\end{definition}
\noindent The definition indicates that a valid IV causally influences the outcome only through the treatment.
As shown in a causal graph of a valid IV ($Z$) in Figure \ref{fig:IV1}, the first condition requires that there is an edge $Z\rightarrow T$ or a non-empty set of collider(s) $\mathbf{X}$ s.t. $Z\rightarrow T \leftarrow \mathbf{X}$, where $\mathbf{X}$ denotes the observed confounders.
The second condition requires that $Z\rightarrow T\rightarrow Y$ is the only path that starts from $Z$ and ends at $Y$. Therefore, blocking $T$ renders $Z\independent Y$.
This implies the \textit{exclusive restriction} that there is no direct edge $Z\rightarrow Y$ or path $Z\rightarrow \mathbf{X}'\rightarrow Y$ where $\mathbf{X}' \subseteq \mathbf{X}$.
Formally, $Y(do(Z=z),T)=Y(do(Z=z'),T)\ \forall T, z\not=z'$.
We denote a dataset with IV as $(\mathbf{x}_i,t_i,y_i,z_i)_{i=1}^N$ where $z_i$ represents the IV of the $i$-th instance. We can then evaluate approaches that leverage IV to estimate ATE, e.g., the ratio estimator~\cite{angrist1992effect}. Exemplar datasets are introduced below.

\begin{itemize}[leftmargin=*]
    \item \textit{1980 Census Extract.} This dataset contains 329,509 observations on the following variables: log weekly wage, quarter of birth (1-4), year of birth (30-39), place of birth (1980 census state codes) and education (highest grade completed).
The goal is to study the causal effect of education on earning.
The quarter of birth is considered as a legitimate
IV~\cite{angrist1995two} for years of schooling. This can be justified by considering that children born earlier in the year enter school at an older age and are allowed to drop out (on their $16$-th or $17$-th birthday)
after having completed less schooling than those born later
in the year.

\item \textit{Current Population Survey Extract.} This dataset contains 30,967 instances. Each instance represents a male born in 1944-53 extracted from the 1979 and 1981-85 March Current Population Survey.
Each instance is matched with a dummy variable which takes value from 25 lottery numbers. There are in total 72 variables.
A potentially valid IV is the lottery number that represents the likelihood of serving in the military during the Vietnam era. This is because serving the military influences both schooling and earnings of a male and is determined by the lottery number. In addition, the lottery number is randomly assigned to each individual.

\item \textit{Slave Export and Trust in Community and Society.} This study evaluates the long-term causal effect of slave export on trust in the community and society~\cite{nunn2011slave}.
In~\cite{puli2019generalized}, the treatment is defined as $t=\ln(1+\frac{\text{slave-export}}{\text{area}})$ and the outcome is the trust in neighbors. The dataset has 6,932 instances and 59 features.
The IV used in~\cite{nunn2011slave,puli2019generalized} is each community's distance to the sea.
\end{itemize}

\subsubsection{Datasets for Regression Discontinuity Design} 
RDD is a type of natural experiment which has been found useful in a variety of real-world problems~\cite{campbell1969reforms}.
A dataset for RDD is denoted by $(\mathbf{x}_i,t_i,y_i,r_i)_{i=1}^N$ where $r_i$ is the running variable of the $i$-th instance that determines the treatment assignment of $i$ based on a predefined cut-off value $r_0$. Running variable is used to control for confounding bias.
For example, in the study of the causal effect of drinking alcohol on youth mortality~\cite{carpenter2009effect}, age is a natural running variable and $r_0=21 \text{(years old)}$.
Here is a list of datasets for RDD methods.
\begin{itemize}[leftmargin=*]
    \item \textit{Population Threshold Dataset~\cite{eggers2018regression}.} Decision-making on many features of municipal governments (e.g., the electoral system, mayors' salaries, and the number of councillors) typically depends on whether the municipality is above or below arbitrary population thresholds. This dataset enables us to evaluate RDD methods for estimating ATEs of threshold-based policies on population-level political and economic outcomes.
    \item \textit{Pretest Scores and Posttest Scores~\cite{jacob2012practical}.} It is used to study the causal effect of students' pre-test scores on their post-test scores.
    This semi-synthetic dataset is generated based on actual student test scores and demographic information. It has 2,767 observations and uses pre-test score as the running variable.
\end{itemize}
\subsubsection{Datasets for Time Series Effect Estimation}
Here list some commonly used real-world datasets for the time series effect estimation. For a more comprehensive list of datasets, please refer to \cite{moraffah2021causal}.
\begin{itemize}
    \item \textit{MIMIC II/III Data} \cite{johnson2016mimic}. It is a large-scale data containing rich information relating to patients admitted to critical care units at a large tertiary care hospital. Some key covariates included are blood pressure, oxygen saturation, given medicine, and other temporal attributes such as time-stamped nurse-verified physiological measurements. 
    \item \textit{Advertisement Data} \cite{brodersen2015inferring}. It is used to measure the effect of an advertisement campaign on the number of times a user was directed to the advertiser’s website from the Google search results page. Particularly, the product-related ads displayed alongside Google's search results for specific key words went live for 6 consecutive weeks. The outcome variable includes search-related visits to the advertiser’s website, such as organic clicks.
    \item \textit{Air Quality Data} \cite{auffhammer2011clearing}. It is used to study whether US gasoline content regulations successfully reduced ozone pollution. The data consists of two sources: data on ambient air concentrations of ozone from the EPA (Environmental Protection Agency)'s Air Quality Standards database for 1989-2003; and weather data measurements from the National Climatic Data Center's Cooperative Station Data (NOAA 2008), such as minimum and maximum temperatures.
\end{itemize}
\subsubsection{Networked Data for Effect Estimation} Here, we present a list of existing semi-synthetic datasets for evaluating causal effect estimation with networked data~\cite{guo2019learning,ma2021causal}.
\begin{itemize}
    \item BlogCatalog~\cite{guo2019learning} is a semi-synthetic observational network dataset. It is based on real-world network structure and node attributes collected from the blog website BlogCatalog. Each node is a blogger and the node attributes are bag-of-words representation of her/his keywords. Each edge represents a friendship relation. The treatments and outcomes are sampled from a set of predefined structural equations, where treatment (control) means the blogger's content is viewed more by mobile (desktop) devices. Outcome is a real number standing for users' opinion on the bloggers' content. Flickr~\cite{guo2019learning} is created in a similar way. Note that these two datasets did not consider interference.
    \item Wave 1~\cite{ma2021causal} is a in-school questionnaire dataset collected by~\cite{chantala1999national}. Authors of~\cite{ma2021causal} creates a KNN graph based on similarity between instances in the original Wave 1 data. The KNN network has 5,578 nodes and 100,158 edges. Each node is a student and node attributes include age, grade, health insurance and so on. Outcome is students' performance. Treatment is whether a student is assigned to a tutor program. Note that this dataset includes simulated peer effect to consider interference.
    \item Pokec~\cite{ma2021causal} is semi-synthetic and is generated from a real-world social network dataset~\cite{takac2012data}. Each node is a user and the node attributes include age, gender, education and so on. Treatment means under exposure of a certain medicine advertisement and outcome represents whether the user made a purchase of it. The purchase can be caused by either the treatment or the interfernce. Similar to Wave 1, the peer effect is simulated from a predefined structural equation.
    
\end{itemize}

\subsection{Discussion}
Revisiting current evaluation pipelines for causal effect estimation manifests that there still remains much to be done towards an effective benchmarking framework. The major challenge we confront when collecting data for causal effect estimation is how to design ethical, fast, reliable, and easy-to-implement experiments. RCTs are mostly impractical given the financial and ethical considerations. We also have limited access to the background variables useful for controlling for confounding bias. For example, users' social network information -- a common hidden confounder in observational studies -- can be used to approximate their socio-economic status~\cite{guo2019learning,guo2019counterfactual}. However, collecting personal data from social networking sites may contradict the terms of user privacy \cite{cheng2021socially}. In terms of effect estimation with networked data (e.g., social network information), in addition to the limited data challenge, extra effort is needed to examine the model robustness against the interference-related assumptions. Given that many outcomes of interest in the social and health sciences might be ordinal and do not have a meaningful scale, model evaluation requires careful treatment in theses scenarios \cite{volfovsky2015causal}. Moreover, collecting data with ground-truth ITEs is almost impossible in reality. This is because counterfactual outcomes cannot be observed at the individual level, otherwise, ITEs would not be needed. In time series causal effect estimation, data collection is even more challenging because outcomes and time-varying covariates may take months or even years to observe \cite{cheng2021long,moraffah2021causal}.

Popular alternatives use synthetic or semi-synthetic datasets where treatments and outcomes are synthesized based on real-world data. Therefore, it is a crucial first step to develop high-quality simulation that mimic the real-world data generating processes for benchmarking causal inference algorithms. From the data perspective, comparatively less effort has been focused on data with auxiliary information under unconfoundedness assumption. Real-world data have rich auxiliary information (e.g., social media data) from multiple modalities and can overcome the limitations in using synthetic and semi-synthetic datasets. Developing and evaluating causal models using such data is important. Another critical missing element in current evaluation procedures is how to validate models with observational data. There have been debates about whether cross-validation -- the standard recipe for validating ML models -- can be used for causal inference. Recently, it is more commonly agreed that it is generally impossible to extend cross-validation in causal inference tasks because cross-validation is used to improve predictive accuracy and assure models' repeatability, but says nothing about causality. Consequently, evaluation metrics and procedures for hyperparameter tuning and causal model selection are in great need. The overarching goal is to develop an open-source platform/software that provides objective, transparent, independent, and easy-to-use evaluation of causal algorithms.
\section{Benchmarking Causal Structure Learning}
Causal structure learning refers to the task of identifying the causal relations for a given set of variables $V = \{X_1, \ldots, X_n\}$. The goal is to generate a causal graph $G=\{V,E\}$ that represents the causal relations over the set of variables in $V$. $E$ represents the set of directed edges between the variables in $V$. Causal relation between two variables $X_i$ and $X_j$ is defined as:
\begin{definition}[Causal Relation]
\textit{For any directed edge $e$ in $G$, $X_i \rightarrow X_j$ implies that $X_i$ is a direct cause of $X_j$ relative to variables in $V$.}
\end{definition}
Causal structure learning can then be defined as
\begin{definition}[Learning Causal Structure]
	\textit{Given observed samples of $J$ variables, $\{X_{,j}\}_{j=1}^J$, we aim to determine whether the value of the $j$-th variable $X_{,j}$ would change significantly if we modify the value of the $j'$-th variable $X_{,j'}$ for all $j\not = j'$.}
\end{definition}
\noindent For instance, learning the causal structure in previous graduate admission example might help answer the question ``\textit{does changing the sex of an applicant causally affect her/his application result?}''. One can use causal graphs to identify the effect that would occur in the other variables when a variable $X_i$ is introduced to an intervention, i.e. when the value of $X_i$ is set to a fixed value. Once a variable is intervened on, a subgraph is generated by removing all the directed edges that point towards $X_i$ in $G$. By repeating this process for all the variables in $V$, one can identify those causal pathways that are active in any experiment.
Given observational data, algorithms for causal structure learning aim to discover a set of causal graphs as candidates~\cite{spirtes2000causation} under the assumption that causality can be identified amongst statistical dependencies~\cite{scholkopf2012causal,peters2017elements}. Methods that recover causal dependence from time-series data have the benefit of temporal precedence and are often based on bivariate Granger causality tests~\cite{granger1969investigating}, which are used to determine whether $X$ causes $Y$ or $Y$ causes $X$ (conditioned on a set of covariates).  Graphical Granger methods use a series of Granger causality tests to determine the full structure among a set of variables.
\subsection{Evaluation Metrics}
\begin{figure}
		\centering
\begin{tikzpicture}
\node[state] (d) at (0,0) {$X_{,2}$};

\node[state] (z) at (2,0) {$X_{,1}$};

\node[state] (y) at (4,0) {$X_{,3}$};

\path (d) edge (z);
\path (z) edge (y);

\end{tikzpicture}
\caption{A different causal graph with the same conditional independence $X_{,2}\independent X_{,3}|X_{,1}$}
\label{fig:equi2}
\end{figure}
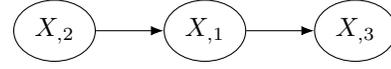

Evaluating causal structure learning compares each of the learned causal graph candidates $\hat{G}$ with a ground-truth $G$ to determine if they belong to the same equivalence class.
\begin{definition}[Equivalence Class]
	\textit{Two causal graphs $G$ and $\hat{G}$ belong to the same equivalence class iff each conditional independence of $G$ is also implied by $\hat{G}$ and vise versa.}
\end{definition}
\noindent For example, Figures~\ref{fig:equi1} and~\ref{fig:equi2} show two causal graphs that belong to the same equivalence class as they share the same set of conditional independence $\left\{X_{,2}\independent X_{,3}|X_{,1}\right\}$.

\begin{figure}[t]
	\centering
		\begin{tikzpicture}

		\node[state] (d) at (0,0) {$X_{,2}$};

		\node[state] (z) at (2,0) {$X_{,1}$};
		
		\node[state] (y) at (4,0) {$X_{,3}$};
		
		\path (z) edge (d);
		\path (z) edge (y);

		\end{tikzpicture}
		\caption{A causal graph with the conditional independence $X_{,2}\independent X_{,3}|X_{,1}$}
		\label{fig:equi1}
		\end{figure}
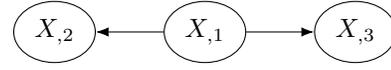

Commonly used evaluation metrics for causal structure learning can be categorized into two major types: (1) graph-distance-based measures and (2) classification-based measures. For the first category,we introduce Structural Hamming Distance (SHD), Structural Intervention Distance (SID), and Frobenius Norm. A comprehensive study of metrics comparing learned causal graphs can be found in~\cite{de2009comparison}. 
\begin{itemize}[leftmargin=*]
    \item \textit{Structural Hamming Distance (SHD)} ~\cite{peters2015structural,tsamardinos2006max,peters2016causal}:
    Given two causal graphs, one being the ground-truth partially DAG\footnote{$\mathcal{G}$ is called a partially DAG if there is no directed cycle, i.e. no pair ($j,k$) such that there are directed paths from $j$ to $k$ and from $k$ to $j$.} and the other being a predicted partially DAG, SHD is defined as the number of edits (adding, removing or reversing an edge) that have to be made to the learned graph $\hat{G}$ for it to become the ground-truth graph $G$. It can be formulated as
\begin{equation}
\label{SHD}
\begin{aligned}
\mathrm{SHD} = A + D + I,
\end{aligned}
\end{equation}
where $A$ represents the number of edges that are added, $D$ represents the number of edges that were deleted, and $I$ represents wrongly oriented edges. Researchers generally present normalized SHD by calculating the ratio of different baselines' SHD over the SHD for their proposed methods.

    \item \textit{Frobenius Norm} measures the difference between two adjacency matrices of two causal graphs \cite{shimizu2011directlingam}. Formally, 
    \begin{equation}
    \label{FrobNorm}
    \text{Frobenius Norm}: \sqrt{\operatorname{trace}\left\{\left(\mathbf{B}_{\text {true}}-\widehat{\mathbf{B}}\right)^{T}\left(\mathbf{B}_{\text {true}}-\widehat{\mathbf{B}}\right)\right\}},
    \end{equation}

where $\mathbf{B}_{\text {true}}$ and $\widehat{\mathbf{B}}$ are the true and predicted adjacency matrices, respectively.
    \item \textit{Structural Intervention Distance (SID)} : For causal structure learning based methods, it is important to understand the causal interpretations of a graph since it helps predict the result of interventions. Given a true DAG $G$ and an estimated DAG $\hat{G}$, SID aims to infer the number of falsely inferred intervention distributions. SID is formulated as~\cite{peters2015structural}
    \begin{equation}
    \label{SID}
    \begin{aligned}
    &\mathrm{SID}: \mathbb{G} \times \mathbb{G} \rightarrow  \mathbb{N}\\
    &(\mathcal{G}, \mathcal{\hat{G}}) \mapsto \#\left\{(i, j), i\neq j\right. \mid \text{the intervention distribution} \\&\text{from $i$ to $j$ is falsely estimated by $\hat{G}$ with respect to $G$}\},
    \end{aligned}
    \end{equation}
    where $\mathbb{G}$ is the space of DAGs defined over variables in $V$. SID is well-suited to evaluate graphs for interventions.
    
\end{itemize}
The second type of metrics are based on the intuition that directional adjacency relations can be treated as a binary classification problem. Therefore, a variety of metrics in classification tasks can be used.
\begin{itemize}[leftmargin=*]
    \item \textit{Precision} is the ratio of true positives (TP) over sum of TP and false positives (FP), i.e., $\text {precision}=\frac{T P}{T P+F P}$.
    \item \textit{Recall} is defined as the ratio of TP over sum of TP and false negatives (FN), i.e., $\text { recall }=\frac{T P}{T P+F N}.$
    \item \textit{F1 score} is the harmonic mean of precition and recall of the learned structure as compared to true causal structure.
    \item \textit{False Positive Rate (FPR)} In terms of graph, FPR is defined as the ratio of the edges that are present in the predicted graph $E_M$ but not present in the ground-truth graph $E_{GT}$ over the absolute difference between ground-truth edges and all possible edges. It is formulated as
    \begin{equation}
        FPR = \mathlarger{\sum}_i \frac{e_i}{|E - E_{GT}|},\ e_i \in E_M \backslash E_{GT}.
    \end{equation}
    \item \textit{True Positive Rate (TPR)} In terms of graph, TPR is defined as the ratio of the common edges between the ground-truth and predicted causal graphs over the number of edges in ground-truth graph. Formally,
    \begin{equation}
        TPR = \mathlarger{\sum}_i \frac{e_i}{|E_{GT}|},\ e_i \in E_M \cap E_{GT}.
    \end{equation}
    
    \item \textit{MSE} is defined as the sum of square of difference between the predicted and the ground-truth causal graphs divided by the total number of nodes. It is formulated as
    \begin{equation}
        MSE = \frac{1}{|T|} \sum (T-A)^{2},
    \end{equation}
    where $T$ is the predicted adjacency matrix and $A$ is the ground-truth adjacency matrix.
    \item \textit{Area under ROC curve (AUC)} is the area under the curve of recall versus FPR at different thresholds.
    \item \textit{Areas under the Precision-Recall  and FPR-TPR curves} can also be used~\cite{tsamardinos2006max,bacciu2013efficient} given that accuracy is measured in terms of precision and recall.
\end{itemize}
Finally, recent works including~\cite{gentzel2019case,peyrard2021ladder} highlight that evaluating against structural measures only, may limit the researchers learning from experimental studies of the model's performance, which in turn creates a gap between theory and practice. To overcome this limitation, they suggest several interventional measures that compare the learned distribution to the ground-truth obtained through interventions. Now we list common interventional measures.
\begin{itemize}[leftmargin=*]
    \item \textit{Total Variation Distance (TVD)}  measures the distance between two probability distributions. Under TVD, the quality of an estimated distribution relative to a known distribution can be computed as
    \begin{multline}
    TVD_{P, \hat{P}, T=t}(O)=\frac{1}{2} \sum_{o \in \Omega(O)}|P(O=o \mid do(T=t))-\\
    \hat{P}(O=o \mid do(T=t))|,
\end{multline}
where $P$ denotes the true interventional distribution, $\hat{P}$ denotes the estimated interventional distribution, and $\Omega(O)$ denotes the domain of $O$.
    \item \textit{KL-Divergence} measures how one probability distribution differs from others. Given the estimated interventional distribution $\hat{P}$ and true interventional distribution $P$, the KL-Divergence for continuous distributions can then be defined as
    \begin{equation}
    D_{KL}(P(x) \| \hat{P}(x))=\int_{-\infty}^{\infty} P(x) \ln \frac{P(x)}{\hat{P}(x)}dx.
    \end{equation}
\end{itemize}
As for time-series data, aside from the aforementioned metrics, F-Test is a popular metric used to evaluate causal structure learning approaches. In this approach, with the defined null hypothesis, parameters are estimated for both the restricted and the unrestricted models. An F-statistic is then computed using the Residual Sum of Squares (RSS) of the two series, which is given by:
    \begin{equation}
    \frac{\left(RSS_{R}-RSS_{UR}\right) / p}{RSS_{UR} /(T-2p-1)} \sim F_{p, T-2p-1},
    \end{equation}
    where $T$ is the length of the time series, $p$ is the number of lags, $RSS_{R}$ is the RSS for restricted model and $RSS_{UR}$ is the RSS for the unrestricted model.
For a more detailed review readers can refer to~\cite{gentzel2019case,peyrard2021ladder}.
\subsection{Evaluation Procedures}
The evaluation procedure for algorithms that learn causal relations from observational data often adopts a transductive setting, where the evaluation is based on the structural difference between the learned graph and the ground-truth graph. This is similar to the standard procedure to evaluate a supervised learning algorithm. Aside from evaluating against structural methods only, recent works emphasize the need for evaluating causal structure learning models against interventional measures. Unlike structural measures, interventional measures penalize edge mis-orientation errors proportionally to their effect on the estimation of interventional effect.

\subsection{Benchmark Datasets}
Datasets in this task can be considered in two categories: (a) The first category includes datasets for learning the causal direction between two variables. 
(b) The second category includes datasets for learning the underlying causal graph of a set of observed variables. Ideal datasets are collected from real-world scenarios and annotated by the experts in the corresponding fields to evaluate algorithms for learning causal relations. However, since it is extremely challenging to obtain such real-world ground-truth datasets for causal graphs, synthetic datasets are commonly used for benchmarking purposes. Popular datasets used for learning causal directions include \textit{Tübingen Cause-Effect Pairs}, \textit{Alzheimer's Disease Neuroimaging Initiative (ADNI)} \cite{petersen2010alzheimer}, \textit{AntiCD3/CD28 }\cite{Sachs523}, and \textit{Note} \cite{mitrovic2018causal}. Datasets for learning causal graphs are \textit{LUng CAncer Simple set (LUCAS)}, \textit{LUng CAncer set with Probes (LUCAP)}, and \textit{Random  Chordal Graphs.} We briefly introduce these datasets. 
\subsubsection{Datasets for Learning Causal Directions}  
We detail exemplar datasets for learning causal relations between two variables as follows:
\begin{itemize}[leftmargin=*]
    \item \textit{Tübingen Cause-Effect Pairs} \cite{mooij2016distinguishing}: It consists of real-world samples from cause-effect pairs and is collected across various subject areas.
    \item \textit{ADNI} \cite{petersen2010alzheimer}: Alzheimer's Disease Neuroimaging Initiative consists of 819 subjects who were enrolled at baseline and followed for 12 months using standard cognitive and functional measures typical of clinical trials.
    \item \textit{AntiCD3/CD28 }\cite{Sachs523}: This is a dataset of causal protein-signal networks with 853 instances of multivariable individual-cell data. Each variable stands for a biomolecule.
    \item \textit{Note} \cite{mitrovic2018causal}: This is a time series data for inferring the arrow (direction) of the time.
    \item \textit{Abalone} \cite{Bache+Lichman:2013}: It contains 4,177 samples and each sample has 4 attributes: Sex, Length, Diameter, and Height. 
\end{itemize}

\subsubsection{Datasets for Learning Causal Graphs}  
\begin{itemize}[leftmargin=*]
    \item \textit{LUCAS and LUCAP.}
    \textit{LUCAS} (LUng CAncer Simple set) and \textit{LUCAP} (LUng CAncer set with Probes) consist of data generated by predefined causal graphs with binary variables. 
    \item \textit{Random  Chordal Graphs.} This is a synthetic dataset generated using approach discussed in~\cite{kocaoglu2017cost,shanmugam2015learning}.
    \item \textit{CausalWorld.} The CausalWorld \cite{ahmed2020causalworld} dataset is a comprehensive robotic benchmark dataset for transfer learning. It provides a combinatorial family of tasks with common causal structures and underlying factors, allowing the user to intervene on the causal variables to determine the similarity level of different tasks.
    \item \textit{SynTReN.} SynTReN is a network generator that creates synthetic transcriptional regulatory networks and produces simulated gene expression data that approximates experimental data. Topologies are created by selecting subnetworks from previous regulatory networks. Results in \cite{van2006syntren} show that the simulated topologies are closer to the real biological networks when compared to different random graph models. Several user-definable parameters adjust the complexity of the resulting dataset w.r.t. the structure learning algorithms.
    \item \textit{ Causality 4 Climate.} The Causality 4 Climate (C4C) \cite{runge2020causality} dataset comes from a NeurIPS 2019 competition for causal structure learning on climate time series. The dataset consists of an extensive number of climate model-based time series datasets with known causal ground-truth. These datasets incorporate the main challenges of causal structure learning in climate research.
\end{itemize}
\subsubsection{Causal Structure Learning Datasets for Time Series}
\begin{itemize}[leftmargin=*]
    \item \textit{Temperature Ozone Data}~\cite{schaechtle2013multi,gong2017causal,mooij2016distinguishing}: This dataset consists of ozone and radiation levels, across 72 points in time, and 16 different places. The assumed ground truth is that the radiation has a causal effect on ozone.
    \item \textit{Neural activity Dataset}: This dataset consists of real-time whole-brain imaging to record the neural activity of Caenorhabditis elegans. The dataset consists of 302 neurons and is generally used to identify which neurons are responsible for the movement.
     \item \textit{Traffic Prediction Dataset}~\cite{pan2018hyperst}: This dataset contains four months’ worth of sensor data from Los Angeles, California. 207 sensors are placed for collecting this data. The location of each sensor in the form of GPS coordinates are also included in the dataset.
     \item \textit{US Manufacturing Growth Data} \cite{entner2010causal}: This dataset consists of microeconomic data of growth rates of US manufacturing firms in terms of employment, sales, research \& development (R\&D) expenditure, and operating income, for the years 1973–2004. It can be used to identify the causal variables that affect the growth rate of a firm.
\end{itemize}
\subsection{Discussions}
The transductive setting has been the norm in the literature of learning causal relations~\cite{spirtes2000causation,malinsky2018causal} in part because the conventional setting of learning causal relations is similar to that of supervised learning. However, in observational studies, the ground-truth causal relations may not be used to train the algorithms even during the training phase. In addition, Guo et al. pointed out in~\cite{guo2020survey} that most of the existing algorithms can only discover the causal relations among variables that can be observed in the training data. Hence, it remains an open problem to develop causal inference algorithms that can be generalized to unseen variables, i.e., an inductive evaluation setting. 
Furthermore, the combination of observational and experimental data may provide with unique opportunities to identify a model, that under various assumptions, can extract some true external cause-effect relationships. There is a recent line of research~\cite{lattimore2016causal,lee2018structural} focusing on learning causal relations from the combination of observational data and interventional data with some selected variables. This is promising as it may overcome the limitations of using pure observational or RCT data. To this end, learning causal relations with the combined data can be reduced to an easier problem where the goal is to identify a certain set of interventions.

\section{Evaluation of Causality-Aware ML Tasks}
As a critical ingredient for AI to achieve human-level intelligence, causal inference has found itself in the spotlight of Trustworthy ML and Socially Responsible AI research \cite{cheng2021socially,cheng2021causal}. In this section, we discuss the evaluation pipeline of two representative tasks: causal interpretability and fairness and unbiased interactive ML.
\subsection{Causal Interpretability and Fairness}
The surge of ML algorithms for decision making in critical fields, such as law-making and autonomous cars, has made it necessary for the decisions made by these models to be interpretable by humans ~\cite{lipton2018mythos}. Interpretability in ML is defined as ``model's ability to explain or to present in understandable terms to a human'' \cite{doshi2017towards}.
To ensure the reliability and transparency of such decisions, causal interpretability aims to explain them by answering the counterfactual questions such as ``what decisions would have been made if they had been under alternative situations (e.g., being trained with different inputs \cite{goyal2019counterfactual, goyal2019explaining} or model components \cite{narendra2018explaining, harradon2018causal})?'' 
We refer to this as a model's \textit{causal interpretability}. Therefore, causal interpretable models are human-friendly and aim to answer causal questions.
%
\subsubsection{Problem Statement}
There is no unified definitions for the causal interpretability of an ML model~\cite{moraffah2020causal}.
In this survey, we use a common definition in which a model's causal interpretability is judged by the \emph{counterfactual explanations} it generates for a set of inputs. Note that we skip the model-level counterfactual explanations in this paper due to the lack of proper evaluation metrics and benchmarks for such frameworks in the literature \cite{moraffah2020causal}.

\begin{definition}[Example-Level Counterfactual Explanation]
	\textit{Given an ML model and a predefined label, a counterfactual explanation of an instance is defined as a new instance that is generated by performing minimal changes to an original instance's features, which makes the model to predict the predefined label.}
\end{definition}
\noindent For example, when a person's credit card application fails, she may be interested in explanations that specify what minimal changes can be made to her profile to pass the application. Note that while similar to the definition of adversarial example, the example-level counterfactual is designed to explain the decision of the model rather than attack it.
Generally, example-level counterfactual explanations aim to answer questions such as ``\textit{Why does this model predict a specific label for an instance?}'' or ``\textit{Was the $i$-th feature of the instance which caused the model to predict this label?}''. Such questions can be answered using counterfactual inference~\cite{pearl2018theoretical}: Counterfactual distributions are new type of conditional probabilities (e.g., $P(y_x|x', y')$) that indicate how likely the outcome of an observed instance, i.e., $y'$, would change given $x'$.

Fairness also relates to making models more transparent and interpretable~\cite{lipton2018mythos}. ML frameworks used for decision making in critical domains such as law-making are required to make fair decisions and not discriminate against specific groups of people \cite{chouldechova2017fair}. Several frameworks have tried to combine causal inference with fairness and propose a criterion for making fair decisions \cite{kusner2017counterfactual, wu2019pc}.
\subsubsection{Evaluation Metrics}
We here discuss common evaluation metrics for two primary tasks in causal interpretability -- example-level counterfactual explanation and fairness.

\noindent\textbf{Example-Level Counterfactual explanations.} One of the big challenges of evaluating counterfactual explanations is the lack-of ground truth in traditional ML datasets. 
A commonly used alternative is to gauge the quality of the counterfactual explanations generated by an ML model in terms of metrics that aim to quantify certain characteristics of these counterfactual explanations. Moraffah et al.~\cite{moraffah2020causal} have classified these metrics based on the properties of the explanation they are designed to measure. Particularly, six metrics -- sparsity, interpretability, proximity, speed, diversity, and visual-linguistic metrics -- have been suggested to measure the quality of the generated counterfactual explanations from different perspectives. 

\textit{Sparsity}  metrics aim to measure the amount of perturbation added to the original input in order to transform it to its corresponding counterfactual explanation. The lower the amount of perturbation is, the better is the quality of the counterfactual. Interpretable counterfactual explanations are often close to the training data manifold \cite{looveren2021interpretable}. To gauge the closeness of counterfactuals to the data manifold, \textit{Interpretability} metrics are utilized and are generally computed based on the reconstruction errors of the counterfactuals \cite{looveren2021interpretable}. \textit{Proximity} is another counterfactual evaluation metric which evaluates how close the counterfactual explanations are to the original samples. Unlike \textit{Interpretability} metrics, Proximity is usually calculated by calculating the $l_p$ distance between the original and counterfacutal samples. Since most interpretablity frameworks are designed for real-world applications, counterfactuals generation approaches should be fast. \textit{Speed} metrics measure the speed of counterfatual generation approaches. Another characteristic of high quality counterfactuals is their diversity. This means that counterfactuals generated for a given input should be different. To measure the quality of this counterfactual explanation frameworks, \textit{Diversity} metrics can be used. Finally, to measure the quality of multi-modal counterfactual explanations such as visual-linguistic explanations, one needs to report the high positiveness/negativeness of the visual explanations as well as the consistency of the visual and linguistic explanations.
Table~\ref{table:counteval} summarizes these metrics with corresponding descriptions. For more detailed explanations, please refer to \cite{moraffah2020causal}. 

 \begin{table*}[]
 \centering
\begin{tabular}{|c|m{15em}|m{25em}|}
\hline
\multicolumn{1}{|c|}{} & \textbf{Counterfactual Characteristic } & \textbf{Description}  \\ \hline \hline
\multirow{2}{*}{1}   & \multirow{2}{*}{Sparsity}  & \multirow{2}{25em}{Measures the amount of perturbation used to generate the counterfactual example \cite{looveren2021interpretable, mc2018interpretable}} \\  &  &  \\ \hline
\multirow{2}{*}{2}  & \multirow{2}{*}{Interpretability}  & \multirow{2}{25em}{Measures the closeness of the counterfactual example to data manifold \cite{looveren2021interpretable}}  \\  & &  \\ \hline 3 & Proximity & Measure the similarity of the counterfactual example to original sample \cite{mothilal2020explaining}  \\ \hline 4  & Speed  & Measures the pace of counterfactual generation \cite{looveren2021interpretable} \\ \hline 5 & Diversity  & Measures the diversity of the generated counterfactuals \cite{mothilal2020explaining} \\ \hline
\multirow{2}{*}{6}     & \multirow{2}{*}{\begin{tabular}[c]{@{}l@{}}Visual-Linguistic \\ Counterfactuals\end{tabular}} & Measures the high positeveness/negativeness of the visual explanation\\  &  & Measures the consistency of the linguistic explanation with it's visual counterpart \cite{DBLP:journals/corr/abs-1812-01263} \\ \hline
\end{tabular}
\caption{A summary of evaluation metrics for counterfactual explanations \cite{moraffah2020causal}.}
\label{table:counteval}
\end{table*}
\noindent\textbf{Fairness.}
Evaluation of causal fairness is challenging and typically done via assessing models' performance in reducing biases. Below, we discuss the state-of-the-art causality-based notions for fairness. For more details, please refer to \cite{DBLP:journals/corr/abs-2010-09553}.

Khademi et al. \cite{khademi2019fairness} propose two notations based on causal effect estiamtion, i.e., FACE (Fair on Average Causal Effect) and FACT (Fair on Average Causal Effect on the Treated), under the potential outcome framework \cite{imbens2015causal}. With FACT notion, a classifier is fair iff:
\begin{equation}
\mathbb{E}[Y_i^{(a_1)} - Y_i^{(a_0)}] = 0,
\end{equation}
which is equivalent to calculating ATE. 
FACT metric, on the other hand, is based on average causal treatment effect on the treated group, i.e., ATT. Based on this definition, a classifier is fair iff:
\begin{equation}
\mathbb{E}[Y_i^{(a_1)} - Y_i^{(a_0)} | A^i = a_0] = 0.
\end{equation}
Another mainstream fairness notion proposed by Kusner et al. \cite{kusner2017counterfactual} is coined counterfactual fairness, which states that a classifier is fair if under any background variable $X$ we have:

\begin{equation}
P(y_{a_1} |X=x, A=a_0) = P(y_{a_0} |X=x, A=a_0),
\end{equation}
where $A$ is a set of protected attributes, $Y$ is the outcome, and $X=V \backslash \{A, Y\}$ denotes the remaining variables in the system. Due to the challenge of causal identification using observational data, Wu et al. \cite{wu2019pc} propose path-specific counterfactual fairness (PC-Fairness), a generalized causal fairness notion which covers various causality-based fairness notions by tuning its hyperparameters. A classifier achieves PC-Fairness iff:

\begin{equation}
P(y_{a_1|\pi, a_0|\bar{\pi}} |O) - P(y_{a_0} |O) = 0,
\end{equation}
where $Y$ is the outcome, $\pi$ denotes causal paths, and $O$ is a set of observed variables.

Built on causal graphs, Zhang et. al \cite{zhang2018fairness} suggest counterfactual direct (Ctf-DE), indirect (Ctf-IE), and spurious effect (Ctf-SE) fairness measures. These metrics essentially assess the transmission from cause to effect. They are defined as:
\begin{equation}
\text{Ctf-DE}_{x_0, x_1} (y|x) = P(y_{x_1,W_{x0}} |x) - P(y_{x_0} |x);
\end{equation}
\begin{equation}
\text{Ctf-IE}_{x_0, x_1} (y|x) = P(y_{x_0,W_{x_1}} |x) - P(y_{x_0} |x);
\end{equation}
\begin{equation}
\text{Ctf-SE}_{x_0, x_1} (y|x) = P(y_{x_0} |x_1) - P(y |x_0),
\end{equation}
\noindent where $X = x_1$ is the intervention on $Y = y$, $X = x_0$ is the baseline value of the sensitive attribute, and $W$ indicates the mediator. Despite recent success in utilizing causality to measure fairness, evaluation of such frameworks is still an open problem mainly due to the ``impossibility of fairness'' \cite{friedler2016possibility}: existing fairness notions appear to be internally consistent but often mutually incompatible with each other.
\subsubsection{Evaluation Procedures}
Counterfactual explanations belong to post-hoc interpretability and are generated after a black-box ML model is trained.
The first step of the evalution procedure is to train an ML model on a regular dataset such as Adult-income~\cite{kohavi1996scaling}.
Then, we need to generate counterfactual explanations based on the trained model and the dataset.
Finally, we evaluate the generated counterfactual explanations using existing criteria such as those in Table \ref{table:counteval}.
The standard procedure in existing work (e.g.,  \cite{mothilal2020explaining}) is transductive.
One extension is its inductive counterpart where we check whether a method can generate counterfactual explanations for an unseen instance from a hold-out set~\cite{wachter2017counterfactual}.
\subsubsection{Benchmark Datasets}

To evaluate the causal interpretability of black-box ML models, we need datasets with two components: (1) a dataset for training and testing an ML task, and (2) a source of ground truth that allows us to evaluate the causal interpretability of the generated explanations.
Ideally, this would be a set of ground-truth counterfactual explanations for a given data instance and an ML model.
Such ground truth can be extremely expensive to acquire manually given the large number of data instances and ML models.
Hence, the second component of ground truth is commonly replaced by a set of proximal metrics~\cite{mothilal2020explaining}. These metrics can be conveniently computed given the generated counterfactual explanations and the corresponding instances from the original dataset.
%
%
Below we introduce common datasets used for interpretability. It is worth mentioning that most of these datasets are not specifically designed for evaluating causal interpretability. Therefore, they do not come with the ground truth of counterfactual explanations that capture the causal aspect of the model. However, the evaluation metrics can be leveraged as proximal measures to determine the quality of a generated counterfactual explanation. Counterfactual explanations are model-agnostic, therefore, can work on datasets for most ML tasks. We categorize these datesets as Image, Text, and Tabular data. Below are exemplar datasets in each category.

\noindent\textbf{Image Datasets.}
\begin{itemize}[leftmargin=*]
    \item \emph{ImageNet (ILSVRC)}~\cite{ILSVRC15}
    is an image dataset organized according to the WordNet hierarchy. 
    There are more than 100,000 \emph{synonym sets} or \emph{synset} in WordNet, majority of them are nouns (80,000+).
    The ImageNet team aims to provide on average 1, 000 images to illustrate each synset. 
    Images of each concept are quality-controlled and human-annotated.
    The task is to predict which synsets an image belongs to.

    \item \emph{MNIST} \cite{mnist_handwritten_digit_database}
consists of images of handwritten digits. It has a training set of 60,000 examples and a test set of 10,000 examples.
It is a subset of a larger set available from NIST. The digits have been size-normalized and centered in a fixed-size image.

    \item \emph{PASCAL VOC 2012} \cite{everingham2010pascal}
    contains 12,031 training and validation images and 1,456 test images. Among the 12,031 training images, 2,913 of them have associated ground-truth object
regions split evenly into a segmentation \emph{train-2012} and
\emph{val-2012} sets.
\end{itemize}
\noindent\textbf{Text Datasets.}
\begin{itemize}[leftmargin=*]
    \item \emph{20 Newsgroup Dataset} \cite{20_NEWS} This dataset is a collection of nearly 20,000 news documents and is partitioned (nearly) evenly across 20 different newsgroups.
    %
    This dataset has become increasingly popular for evaluating text mining and natural language processing algorithms w.r.t. text classification and clustering tasks.
    

    \item \emph{IMDB} \cite{imdb} The IMDB dataset contains 25,000 training documents, 25,000 test documents, and 50,000 unlabeled documents. It is a dataset for binary sentiment classification task. The dataset consists of movie reviews retrieved from the website IMDB\footnote{https://www.imdb.com/}. 
    For labeled
documents, they are sampled in a way such that there is a 1:1 ratio between negative
and positive documents.
%
    \item \emph{Amazon} reviews~\cite{amazon} This dataset contains product reviews and metadata from Amazon, including up to 142.8 million reviews spanning May 1996 - July 2014. Particularly, it includes reviews (ratings, text, helpfulness votes), product metadata (descriptions, category information, price, brand, and image features), and links (also viewed/also bought graphs).
One of the tasks can be evaluated by this dataset is binary sentiment classification.
    
\end{itemize}
\noindent\textbf{Tabular Datasets.} The UCI repository~\cite{uci_repository} provides a myriad tabular datasets used by the ML literature. Here, we list those that are widely used for evaluating causal interpretability.
\begin{itemize}[leftmargin=*]
    \item \emph{Adult-income}~\cite{kohavi1996scaling} This dataset contains demographic, educational, and other information based on 1994 Census database and is available on the UCI ML repository~\cite{uci_repository}. This dataset is preprocessed to include 8 features, namely, hours per week, education level, occupation, work class, race, age, marital status, and sex. The ML model’s task is to classify whether an individual’s income is over $50,000$.
    
\item \textit{German loan dataset} \cite{Dua:2019}. This dataset contains 1,000 observations of loan applicants including numeric, categorical, and ordinal attributes. The label indicates whether the application is successful.

\item \textit{LendingClub}\footnote{{https://www.lendingclub.com/info/download-data.action}}.
This dataset consists of 5 years of loan records (2007-2011) by LendingClub company. After the dataset is preprocessed, it contains 8 features: employment years, annual income, number of open credit accounts, credit history, loan grade
as decided by LendingClub, home ownership, purpose, and the state
of residence in the U.S.
\item \textit{COMPAS}. This is a dataset collected by ProPublica~\cite{article12}. It contains information for analysis on recidivism decisions in the U.S. After preprocessing, this dataset contains the following 5 features: bail applicant's age, gender, race, prior count of offenses, and the degree of criminal charge. 

\end{itemize}

\subsection{Unbiased Interactive ML}
Recent years have witnessed increasing importance of interactive ML models and systems such as search ranking and recommendation systems in both academia and industry. Training such interactive models and systems is different from standard prediction tasks because the supervision signals or labels -- e.g., ratings, clicks, and purchases -- come from users' online behavior, collected through interactions between users and an ML model.
The fundamental challenge in learning interactive ML models arises from the fact that they are trained/optimized with biased log data whilst the goal is to optimize the performance via online experiments where RCTs can be conducted~\cite{guo2020debiasing,wang2020causal}.
This mismatch between training and test data makes causality critical in understanding and mitigating various types of bias in the process of training interactive ML models.
In this section, we illustrate the uses of CL in one of the extensively studied tasks in interactive ML -- unbiased learning to rank.
Another related task is debiasing recommendation systems~\cite{chen2020bias}.

\subsubsection{Problem Statement}

Unbiased learning to rank is formally defined as follows.
\begin{definition}[Unbiased Learning to Rank]
Given a search result page (SERP) of a query $q$ in the search log, $\bm{X}^q\in \mathcal{R}^{m^q\times d}$, $\bm{y}^q \in \{1,...,m^q\}^q$, and $\bm{c}^q\in\{0,1\}^q$ denote the feature matrix, the vector of item indices, and the vector of implicit feedback, respectively. $d$ is the number of features and $m^q$ is the number of items shown in the SERP of query $q$.
Let $\mathcal{Q}$ and $\mathcal{Q'}$ be the set of queries for an offline training set and that of a test set, respectively.
$f:\mathcal{R}^{d}\rightarrow \mathcal{R}$ denotes a ranking system that maps the feature vector of an item to its ranking score.
Given offline search log data $\{\bm{X}^q,\bm{y}^q,\bm{c}^q\}_{q\in\mathcal{Q}}$, the goal is to optimize a specific ranking metric (e.g., NDCG) in a test set $\{\bm{X}^q,\bm{r}^q\}_{q\in\mathcal{Q}'}$ where $\bm{r}^q$ denotes the vector of unbiased feedback. $\bm{r}^q$ can be collected from RCTs or manual labeling (e.g., relevance score).
\end{definition}
In an RCT, we rank all the items randomly and collect feedback from users.
We call such feedback \emph{unbiased} as it is only causally influenced by users' preference.
While with offline log data, a user's feedback on an item is influenced by the user's preference, the position of the item ranked by the logging policy -- the existing ranking system that generated the log data, and other potential factors (e.g., context of the item~\cite{wuunbiased}).
In some cases, if we do not have access to unbiased implicit feedback from RCTs for the test set, we can also evaluate a ranking system based on relevance scores as in ~\cite{joachims2018deep,hu2019unbiased,ai2018unbiased}.

\subsubsection{Evaluation Procedures} 
The evaluation procedure of unbiased learning to rank is mostly similar to that of a standard learning to rank task in information retrieval.
The major difference is that the training set and the test set are from different sources: the former is a typically biased offline dataset whereas the latter is a dataset with unbiased feedback (often collected through RCTs).
Given a test set $\{\bm{X}^q,\bm{r}^q\}_{q\in\mathcal{Q}'}$ and the corresponding predicted ranking $\{\hat{\bm{y}}^q\}_{q\in\mathcal{Q}'}$ of a ranking system, we introduce the following two most popular metrics to evaluate its performance.

\noindent\textit{Normalized Discounted Cumulative Gain (NDCG@K)}~\cite{jarvelin2002cumulated} is defined as  \begin{equation}
         \text{NDCG@K} = \frac{1}{\text{IDCG@K}}\sum_{i=1}^K\left(\frac{2^{r^q_i}-1}{\log_2(i+1)}\right),
\end{equation}
where $K$ is a positive integer (e.g., 10 or 50) and varies by specific applications; $\text{IDCG@K} = \sum_{i=1}^K\frac{1}{log_2(i+1)}$ is the normalizer to ensure $\text{NDCG@K}$ is in the range $[0,1]$. Note that an item's rank is determined by the prediction $\hat{\bm{y}}^{q}$ of a ranking system.
NDCG@K is a weighted sum of a function of unbiased feedback (e.g., relevance score) for the items ranked in top-K positions.
A larger NDCG@K indicates that the top-K ranked items are more relevant.

\noindent\textit{Mean Average Precision (MAP@K)} is defined as
\begin{equation}
           \text{MAP@K} = \frac{1}{K}\sum_{i=1}^K{|\{j|r^q_j=1,j=1,...,i\}|}/i.
\end{equation}
MAP@K describes how accurate a ranking system's ranked predictions are, on average, over a whole validation/test dataset. The primary difference between MAP@K and NDCG@K is that the former assumes binary relevance, i.e., either relevant or irrelevant, whereas the latter also takes continuous values.

In addition, there are metrics proposed specifically for debiasing popularity bias of ranking algorithms.
The first metric is \textit{Average Recommended Popularity (ARP)}~\cite{abdollahpouri2019managing}, defined as:
\begin{equation}
    \text{ARP@K} = \frac{1}{|\mathcal{Q}'|}\sum_{q\in\mathcal{Q}'} \frac{\sum_{i=1}^K \phi_i^q}{K},
\end{equation}
where $\phi_i^q$ denotes the popularity of the document or product ranked at position $i$ in the SERP of query $q$.
We can also focus on the documents or products that are least popular, namely the long-tail documents or products.
This leads to the \textit{Average Percentage of Long Tail Items (APLT@K)}~\cite{abdollahpouri2019managing}, defined as:
\begin{equation}
    \text{APLT@K} = \frac{1}{|\mathcal{Q}'|}\sum_{q\in\mathcal{Q}'} \frac{\sum_{i=1}^K1(i\in\mathcal{LT})}{K},
\end{equation}
where $1(\cdot)$ is an indicator function. $\mathcal{LT}$ denotes a set of long-tail documents or products pre-defined by their popularity.
\subsubsection{Benchmark Datasets}
Data collection in unbiased learning to rank confronts general challenges in CL tasks. The alternative seeks to create semi-synthetic datasets using a search log dataset with ground-truth relevance scores for each query-item pairs~\cite{ai2018unbiased,hu2019unbiased}.
First, we train a ranking system (e.g., RankSVM ~\cite{ai2018unbiased}) on 1\% of the original training dataset. We then generate a list of items ranked in order using the entire training dataset $\{\bm{X}^q,\bm{r}^q\}_{q\in\mathcal{Q}}$.
The next step employs a click model (e.g., the position-based click model~\cite{craswell2008experimental}) to simulate users' clicks.
%
%
Specifically, the position-based model assumes $P(c_i^q=1) = P(o_i^q=1)P(z_i^q=1)$, indicating that a click on $i$-th item can only be observed iff the item is examined ($o_i^q = 1$) and relevant ($z_i^q=1$).
The binary relevance label $z_i^q$ is defined as a function of the relevance score $r_i^q$:
\begin{equation}
    P(z_i^q=1) = \epsilon +   (1-\epsilon) \frac{2^{r_i^q}}{2^{4}-1},
\end{equation}
where $\epsilon$ denotes the click noise and is often set to 0.1~\cite{ai2018unbiased,hu2019unbiased}. The maximum value of $r_i^q$ is $4$.
The probability of examination is defined as a function of the position-based examination probability $\bm{p}$~\cite{ai2018unbiased,hu2019unbiased}:
\begin{equation}
    P(o_i^q = 1) = (p_i)^\eta,
\end{equation}
where $\eta\ge 1$ is a hyperparameter set to $1$ by default.
One popular dataset used for evaluating unbiased learning to rank models is the \textit{Yahoo! Learning to Rank Challenge Dataset}. It contains 29,921 queries and 710k items~\cite{chapelle2011yahoo}. Each query-item pair is represented by a feature vector of 700 dimensions and associated with a ground-truth relevance score in $\{0,1,2,3,4\}$. Any search log dataset with ground-truth relevance scores or feedback collected through RCTs can be potentially used in this task.

\section{Causal Inference Tools}
In this section, we examine popular tools/packages for benchmarking causal inference, including tools for causal effect estimation -- \textit{CausalML, EconML, DoWhy CauseBox}, for causal structure learning -- \textit{CausalNex, CausalDiscovery, pcalg, bnlearn, TETRAD}, and for evaluation -- \textit{Causality-Benchmark, JustCause}.
\begin{itemize}[leftmargin=*]
    \item \textit{CausalML} \cite{chen2020causalml} implements an array of uplift modeling and causal inference methods using ML algorithms. It provides a standard interface for users to estimate CATE or ITE based on experimental or observational data, without strong assumptions on the model form. CausalML currently supports Tree-based algorithms (e.g., Uplift random forests on KL divergence), Meta-learner algorithms (e.g., S-learner, T-learner), and IV algorithms (e.g., 2-Stage Lease Squares). Covered evaluation metrics include RMSE and MAE.
    \item \textit{EconML} \cite{econml} estimates heterogeneous treatment effects from observational data via ML. The goal of EconML is to combine state-of-the-art ML techniques with econometrics to automatically solve complex causal inference problems. The supported estimation methods are in the intersection of econometrics and ML, including double ML, orthogonal random forests, meta-learners, doubly robust learners, orthogonal IV, and deep IV. It does not include evaluation metrics for effect estimation.
    \item Informed by conventional ML libraries for prediction, \textit{DoWhy} \cite{dowhy} provides a unified interface for causal inference methods under the two fundamental frameworks -- graphical models and potential outcomes. First, DoWhy creates a causal graphical model for each problem to describe the causal assumptions. It then uses graph-based criteria and \textit{do}-calculus to find all potential ways of identifying a desired causal effect. In the third step, it estimates causal effects based on the identified estimand. Finally, DoWhy validates an effect estimate from a causal estimator using multiple refutation methods such as adding unobserved common causes and bootstrap validation. DoWhy can also call external estimation methods such as EconML and CausalML. 
    \item \textit{CausalNex}\footnote{https://github.com/quantumblacklabs/causalnex} uses Bayesian Networks to combine ML and domain expertise for causal structure learning and effects estimation. It deploys state-of-the-art structure learning method DAGs with NO TEARS \cite{zheng2018dags} to understand the causal relationships between variables. CausalNex allows users to learn optimal graph structure through: 1) encoding domain expertise and 2) learning from data via structure learning algorithms. Users can also conduct counterfactual analysis by introducing \textit{do}-calculus.
    \item \textit{CausalDiscovery} \cite{kalainathan2019causal} implements an end-to-end, step-by-step pipeline approach for recovering the direct dependencies (the skeleton of the causal graph) and the casual relationships between variables. CausalDiscovery currently includes 17 algorithms for graph skeleton identification and 19 algorithms for causal directed graph prediction. The goal of CausalDiscovery is to learn both the causal graph and the associated causal mechanisms from the joint probability distribution of observational data. It also includes R based algorithms to extend the toolkit with additional R packages. There are two types of supported algorithms in CausalDiscovery -- graph and pairwise causal inference models. In addition to the standard evaluation metrics, CausalDiscovery also includes SHD and SID.
    \item \textit{Causality-Benchmark} \cite{2018_CausalBenchmark} benchmarks algorithms that estimate ATE and ITE of an intervention on the outcome of interest. It includes unlabeled data for prediction, labeled data for validation, and scoring algorithms for automatic evaluation of prediction algorithms based on different evaluation metrics. As ground-truth data of causal effect cannot be known for real-world treatment, Causality-Benchmark uses a simulation-based approach that leverages existing covariates and creates a causal graph to determine treatment assignment and effect. It supports a variety of metrics for causal effect estimation.
    \item \textit{JustCause}\footnote{https://github.com/inovex/justcause} evaluates causal inference methods using common datasets including Twins, Infant Health and Development Program (IHDP), and IBM ACIC. It can also generate synthetic datasets with a generic but standardized approach used in \cite{wager2017estimation}. The goal of JustCause is to provide a fair and just way to benchmark new methods against several baselines and the state-of-the-art methods in causal effect estimation. The supported algorithms are doubly robust estimation, inverse propensity weighting, S-learner, and T-learner. Evaluation metrics implemented include PEHE score, MAE, EnoRMSE score, and Bias.
    \item \textit{CauseBox}\footnote{https://github.com/paras2612/CauseBox} is a benchmarking suite developed in Python for treatment effect estimation methods that consists of an ensemble of five deep learning based SOTA methods and two tree based methods. All the methods are evaluated on benchmarking datasets including IHDP and News datasets. The goal of this toolbox is to provide a unified platform to compare algorithms for different metrics including PEHE and Policy Risk.
    \item \textit{bnlearn}\footnote{https://www.bnlearn.com/}~\cite{scutari2010learning} is a causal structure learning toolbox developed in the R language. The bnlearn package provides implementations of various structure learning algorithms including, but not limited to, PC; Grow-Shrink (GS); Incremental Association Markov Blanket (IAMB); Hybrid Parents \& Children (HPC); and Hill Climbing (HC). Bnlearn also provides the conditional independence tests and network scores used to construct the Bayesian network. Both discrete and continuous data are supported. Furthermore, bnlearn facilitates choosing the learning algorithms based on different statistical criteria, so that the best combination for the data could be utilized.
    \item \textit{TETRAD}\footnote{http://www.phil.cmu.edu/tetrad/}~\cite{ramsey2018tetrad} is a drag and drop suite to perform causal structure learning. It can take datasets containing both continuous and categorical variables, including time series data. It supports algorithms to search for structural relations. It also supports measuring unmeasured confounders; simulating data from a statistical model; predicting the effects on other variables of interventions or perturbations on one or more variables; and computing the probability distribution of any variable conditional on specified values of any other set of variables. It is developed in Java.
    \item \textit{pcalg}\footnote{https://cran.r-project.org/web/packages/pcalg/index.html}~\cite{kalisch2012causal} is a toolbox developed in R for causal structure learning and causal effect estimation using graphical models. For structural learning, pcalg supports multiple algorithms, including PC, FCI, RFCI, and GIES, whereas for causal effect estimation, it includes the IDA algorithm, the Generalized Backdoor Criterion (GBC), and the Generalized Adjustment Criterion (GAC).
    
\end{itemize}
Comparisons of these tools w.r.t. data, supported methods, and metrics can be found in Table \ref{Tools}. An important notion is that data types supported by most of these tools are limited to i.i.d. data and data with IV. Further, current evaluation-oriented tools do not support metrics for heterogeneous effect estimation and for causal structure learning tasks. In addition, none of these tools are designed to evaluate causality-aware ML tasks, e.g., causal interpretability.

\begin{table*}
\resizebox{\textwidth}{!}{
\begin{tabular}{|c|c|c|c|c|c|c|c|c|c|c|c|c|}
\hline
\multicolumn{2}{|c|}{Task}                         & \multicolumn{4}{c|}{Causal Effect Estimation}         & \multicolumn{5}{c|}{Causal Structure Learning} & \multicolumn{2}{c|}{Evaluation for Effect Estimation} \\ \hline
\multicolumn{2}{|c|}{Tool}                         & CausalML & EconML                               & DoWhy &CauseBox & CausalNex &pcalg &bnlearn &TETRAD    & CausalDiscovery      & Causality-Benchmark             & JustCause             \\ \hline
\multirow{4}{*}{Data}     & i.i.d                    & \cmark        & \cmark                                    & \cmark    &\cmark  & \cmark &\cmark &\cmark &\cmark             & \cmark                    &      \cmark                    & \cmark                     \\ \cline{2-13} 
                          & IV                     & \cmark        & \cmark                                    & \cmark     &  & &\cmark &\cmark &\cmark   &                      &                                 &                       \\ \cline{2-13} 
                          & Networked              &          &    &                                  &       &  & & &                &                      &                                 &                       \\ \cline{2-13} 
                          & Time Series            &          &     &                                 &    & & & &\cmark                  &                      &                                 &                       \\ \hline
\multirow{9}{*}{Methods}  & Propensity Score       &      &    & \cmark                                    & \cmark     &    & & &              &                      &                                 & \cmark                     \\ \cline{2-13} 
                          & Tree-based    & \cmark        & \cmark                                    & \cmark     &\cmark &         & &\cmark &        &                      &                                 &                       \\ \cline{2-13} 
                          & Meta-Learner           & \cmark        & \cmark                                    & \cmark     &  &       & & &         &                      &                                 & \cmark                     \\ \cline{2-13} 
                          & Doubly ML              &          & \cmark                                    & \cmark     &  &      & & &            &                      &                                 &                       \\ \cline{2-13} 
                          & Doubly Robust &          & \cmark                                    & \cmark     &  &              &\cmark    & & &                   &                                 & \cmark                     \\ \cline{2-13} 
                          & IV                     & \cmark        & \cmark                                    & \cmark     &  & & & &              &                      &                                 &                       \\ \cline{2-13} 
                          & Mediation              &  &         &                                      &     &    & & &              &                      &                                 &                       \\ \cline{2-13}
                          & Graph            & &          &                                      &       & \cmark          &\cmark &\cmark &\cmark      & \cmark                    &                                 &                       \\ \cline{2-13} 
                          & Pairwise               &          &                                      &        &\cmark &   & & &               & \cmark                    &                                 &                       \\ \hline
\multirow{11}{*}{Metrics} & PEHE                   &          & \multirow{11}{*}{User-Input Metrics} &       &\cmark  & & & &                &                      &                                 & \cmark                     \\ \cline{2-3} \cline{5-13} 
                          & RMSE                   & \cmark        & &\cmark                                      &\cmark       & & &\cmark &                 &\cmark                      & \cmark                               & \cmark                     \\ \cline{2-3} \cline{5-13} 
                          & MAE                    & \cmark        &  &\cmark                                    &\cmark       & & & &                &                      &                                 & \cmark                     \\ \cline{2-3} \cline{5-13} 
                          & Bias                   &          &      &                                &\cmark       &   & & &               &                      & \cmark                               & \cmark                     \\ \cline{2-3} \cline{5-13} 
                          & Coverage               &          &   &                                   &       &  & & &                &                      & \cmark                               &                       \\ \cline{2-3} \cline{5-13} 
                          & Confidence Interval    &          &  &                                    &       &      & & &            &                      & \cmark                               &                       \\ \cline{2-3} \cline{5-13} 
                          & Aggregating Score      &          &   &                                   &       &        & & &          &                      & \cmark                               &                       \\ \cline{2-3} \cline{5-13} 
                          & Refutation             &          &    &                                  &     &    & & &             &                      &                                 &                       \\ \cline{2-3} \cline{5-13} 
                          & SID                    &          &      &                                &       &    & & &              & \cmark                    &                                 &                       \\ \cline{2-3} \cline{5-13} 
                          & SHD                    &          &      &                                &       &    &\cmark &\cmark &\cmark              & \cmark                    &                                 &                       \\ \cline{2-3} \cline{5-13} 
                          & Classification         &          &      &                                &   &\cmark &\cmark &\cmark      & \cmark              & \cmark                    &                                 &                       \\ \hline
\end{tabular}
}
\caption{Comparisons of causal inference tools with a focus on the included datasets, methods, and metrics.}
\label{Tools}
\end{table*}
\section{Open Problems and Challenges}
Still an early-stage research in AI, CL presents both great opportunities and multi-faceted challenges. We list a few to bring to the front some major concerns in the field.
\subsection{Evaluation of Intermediate Steps} The importance of validating intermediate steps in a CL pipeline is often overlooked. For example, it is critical to precisely infer propensity scores as it is an intermediate step in many popular causal methods, e.g., propensity score matching. Therefore, an evaluation pipeline that effectively evaluates intermediate steps is desired. In text analysis, as text is interpretable and can be understood by humans, interpretable balance metrics such as standardized difference in means and/or human judgements of propensity scores can be used to assess the predicted propensity scores or matching results \cite{keith2020text}. How to improve and standardize human judgement experiments, however, remains an open problem. 
\subsection{Evaluation for Networked Data}
Networked data (e.g., social network) have been of great interest to researchers due to its ubiquity in real world. Some representative tasks include community detection and predicting relationships between individuals in the network. Despite the growing interest in identifying and estimating causal effects in networked data, methodologies and their evaluation have not kept apace \cite{ogburn2017causal}. Due to the interference and peer effects between units, standard Structural Causal Models are not equipped to deal with dependence across individuals and describe the data generating process of networked data. This further impedes the research in causal structure learning for networked data.
On the positive side, many evaluation metrics for causal inference and CL with i.i.d. data can still be used to benchmark methods for the corresponding tasks with networked data. The major challenge lies in the lack-of real-world datasets. Most existing work (e.g., ~\cite{guo2019learning,bevilacqua2021size}) relies on synthetic data generated from graph models or semi-synthetic data with predefined structural equations. 
\subsection{Constructed Observational Studies} Constructed data -- data gathered from both randomized and non-randomized experiments with similar subjects and settings -- have been used in many areas, such as economics \cite{lalonde1986evaluating}, marketing \cite{gordon2019comparison}, and education \cite{shadish2008can}. AI researchers have also started using constructed data, as evidenced by recent works that leverage the advantages of both RCTs and observational studies to enhance online A/B tests (e.g., \cite{ye2020combining}). Using constructed data for CL makes up for the limitations in using only observational data or RCTs and helps make better decisions. Further, it reveals possible solutions to evaluate intermediate steps and validate the process of model selection and hyperparameters tuning. It also allows us to evaluate causal effect estimation (e.g., ITE) when the ground truth is not available.
\subsection{Evaluation of Long-Term Effects} Short- and long-term effects can differ in a number of ways. For example, user behavior changes as they learn and adapt to new environment. In a social network, user behavior is easily influenced by people in their network though the change may take time to reach its full effect \cite{kohavi2017online}. Different from evaluating short-term effects, evaluating long-term effects is more cumbersome as we need to consider the factor of time. Current evaluation methods directly adopt pipelines in short-term effect estimation, e.g., MAE of the estimated effect at a specific timestep. We need metrics tailored to the unique challenges in the long-term effect estimation. For example, when is the best target inferential point to estimate the effect? The answer reveals how long it takes a user to discover a new feature in a recommender system and if in-product education is needed to expedite the uptake. Many issues may arise given the nature of the data available to investigate long-term effects \cite{newsome2018estimating}. For instance, it is possible that a treatment effect depends on the previous levels of covariates, therefore, evaluation measures that can test for the presence and strength of effect modification are preferred. Sensitivity analysis and model validation are especially important in this task due to the issues of mis-specifying the direction of causal pathways and censoring, e.g., people lost to follow-up over time.
\subsection{Model Validation}
Causal inference is generally an unsupervised learning technique, which makes the model validation challenging. With the required (and often unverifiable) causal assumptions, a valid and unbiased causal estimate can only be assured after the sensitivity and robustness of an estimate against these assumptions are checked. For example, despite the promising research progress on addressing unobserved confounding, current approaches are less readily available (or impossible) to remove all confounding. Even for unbiased estimators in the presence of unobserved confounders, assumptions required for causal identification and estimation can be violated. Therefore, being able to determine the robustness of causal methods is crucial for theory testing and model development. In addition, knowing to what extent the conclusions drawn from those causal studies are sensitive to potential confounding and violations of assumptions can help with policy decision making \cite{liu2013introduction}. Although there exists an extensive literature of sensitivity analysis \cite{huang2020model}, most results are limited to specific model structures and on a case-by-case basis, whereas a general-purpose algorithmic framework for sensitivity analysis is still lacking to account for the ubiquity of causal questions in the sciences and AI \cite{cinelli19a}.
\subsection{CL Evaluation Tools} The number of metrics and datasets for CL is fast growing. A pressing need for an open-source platform is thereby raised to promote objective, transparent, and independent evaluation of algorithms and to support broad collaborations. Such a platform is expected to include the ever-growing evaluation metrics, procedures, and datasets in causal effect estimation, causal structure learning, and causality-aware ML tasks. Model validation tools such as sensemakr\footnote{\url{https://cran.r-project.org/web/packages/sensemakr/index.html}} are also in need to better understand the robustness of a causal model. By providing convenient access to the benchmarking repository for evaluating CL algorithms, we can enlarge the collaboration in the community for the development of new metrics and datasets. 
\subsection{Evaluation of Causal Interpretability}
Evaluating interpretable models is particularly difficult since intepretability does not have a unified definition \cite{moraffah2020causal}. Therefore, no specific measure can be used to fully assess models' intepretability from various aspects. It is even more challenging to evaluate models for causal interpretability due to the lack of ground-truth causal relations between the components of the model or the causal effect of one component on another. Therefore, crafting new datasets with such ground truth is a vital task for the field to move forward.
\section{Conclusion and Future Work}
CL is essential for AI to uncover causal mechanisms underlying real-world problems and to achieve human-level intelligence. Still at its infancy stage, research in CL is facing many obstacles. For example, CL has burgeoned, so far, without a proper benchmarking pipeline to support fair and transparent evaluations of emerging research contributions. To bridge this gap, in this survey, we provide a comprehensive review of existing methods for evaluating the fundamental tasks in causal inference and causality-aware ML tasks and discuss potential limitations. We follow the evaluation pipeline in conventional ML to focus on the widely-used datasets, evaluation metrics, protocols, and causal inference tools/packages. We conclude the survey with prominent open problems and challenges that await future research. To this end, we hope to broaden the discussions about open-source platforms/software that aim to promote CL research via fair, transparent, independent, and easy-to-use evaluation procedures.

We plan to develop a Causal Inference Evaluation Toolbox that consists of mainstream CL algorithms such as CounterFactual Regression (CFR) \cite{shalit2017estimating}, Causal Effect Variational Autoencoder \cite{louizos2017causal}, Bayesian Regression Trees (BART) \cite{hahn2017bayesian}, Causal Forest \cite{wager2018estimation}, Perfect Match \cite{schwab2018perfect}, Disentangled Representations for CFR \cite{hassanpour2019learning}, and similarity
preserved individual treatment effect \cite{yao2018representation}. This toolbox will showcase results of previously mentioned evaluation metrics for multiple benchmark datasets such as Jobs, IHDP, and News dataset. 

\section*{Acknowledgements} This work is supported by National Science Foundation (NSF) grants \#1909555, \#2029044, \#2125246, \#1633381, \#1610282, ARL W911NF2020124, and AMC W911NF2110030. The views, opinions, and/or findings expressed are those of the authors and should not be interpreted as representing the official views or policies of the Army Research Office or the U.S. Government.
\bibliographystyle{IEEEtran}
\bibliography{Reference}

\begin{IEEEbiography}[{\includegraphics[height=1.25in,clip,keepaspectratio]{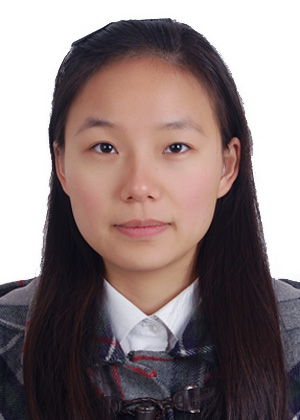}}]{Lu Cheng} received her B.Eng. degree in Logistic \& Systems Engineering from Huazhong University of Science and Technology and M.Eng. in Industrial Engineering from Rensselaer Polytechnic Institute. She is currently a fifth-year PhD candidate of Computer Science and Engineering at Arizona State University. Her research interests include socially responsible AI, CL, and data mining. She has published research papers in premier conferences of AI, data mining and ML. She is a student member of the ACM. Contact her at lcheng35@asu.edu.
\end{IEEEbiography}

\begin{IEEEbiography}[{\includegraphics[height=1.25in,clip,keepaspectratio]{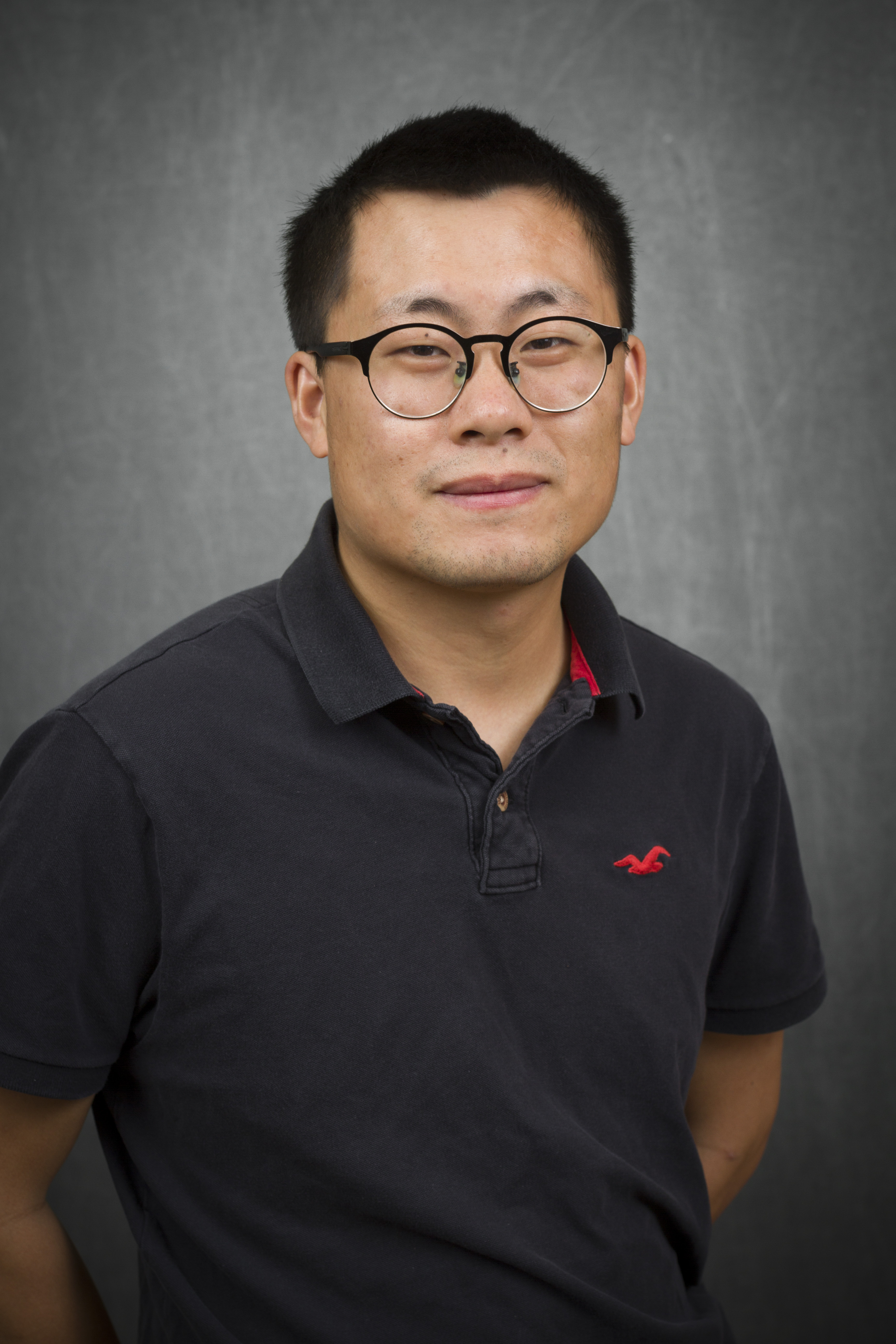}}]{Ruocheng Guo} received his Ph.D. degree in Computer Engineering from Arizona State University. He is currently an Assistant Professor of Data Science at City University of Hong Kong. His research interests include causal ML towards fair, interpretable and generalizable AI, causal inference and data mining. He is a member of the ACM, SIAM, and AAAI. Contact him at ruocheng.guo@cityu.edu.hk.
\end{IEEEbiography}

\begin{IEEEbiography}[{\includegraphics[height=1.25in, width=1in,clip,keepaspectratio]{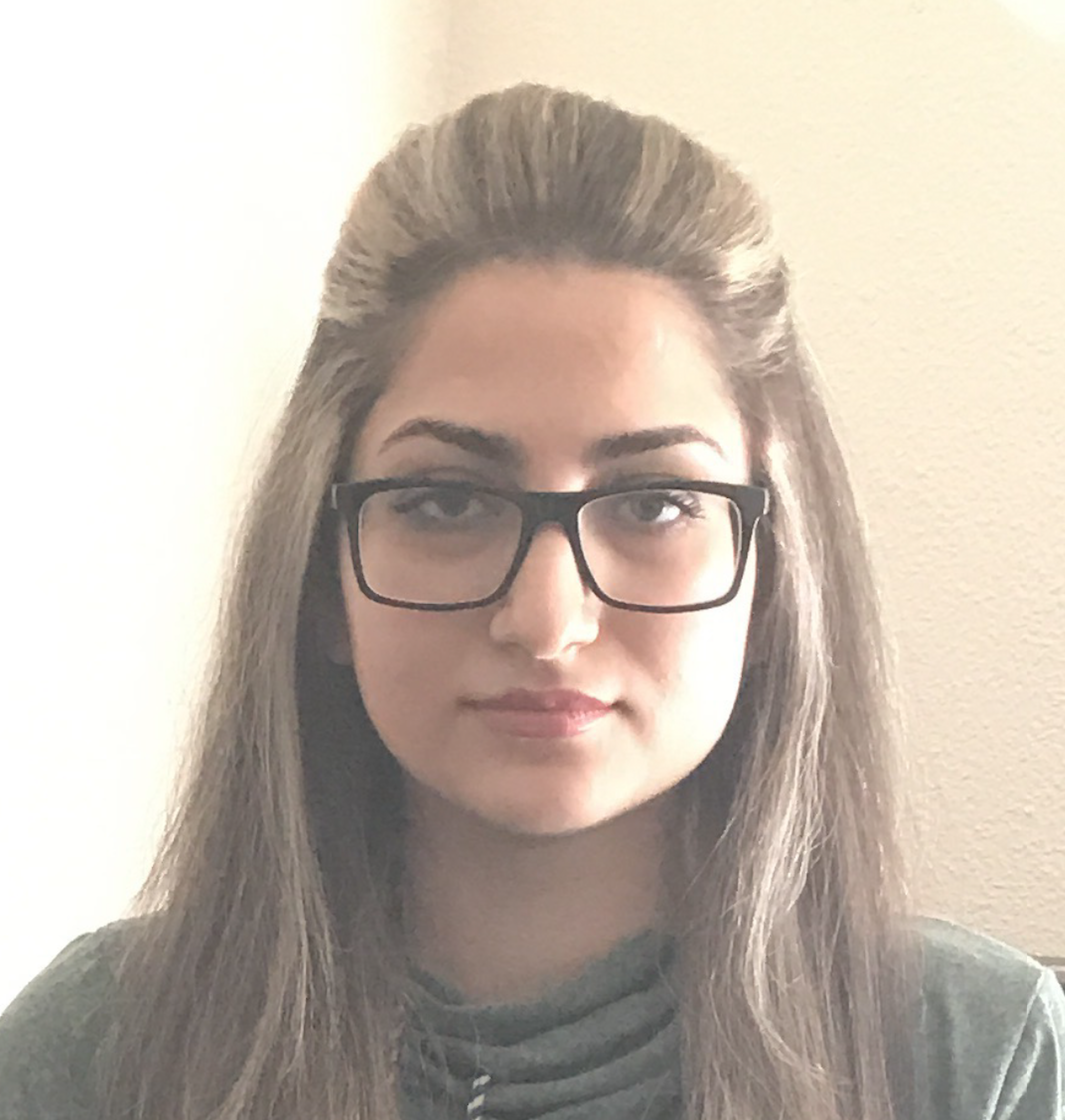}}]{Raha Moraffah} received her B.S. degree in Computer Sience and Engineering from Sharif University of Technology. She is currently a fifth-year PhD student of Computer Science and Engineering at Arizona State University. Her research interests include causal inference, causal ML and adversarial learning. She is a student member of ACM. Contact her at rmoraffa@asu.edu.
\end{IEEEbiography}

\begin{IEEEbiography}[{\includegraphics[height=1.25in, width=1in,clip,keepaspectratio]{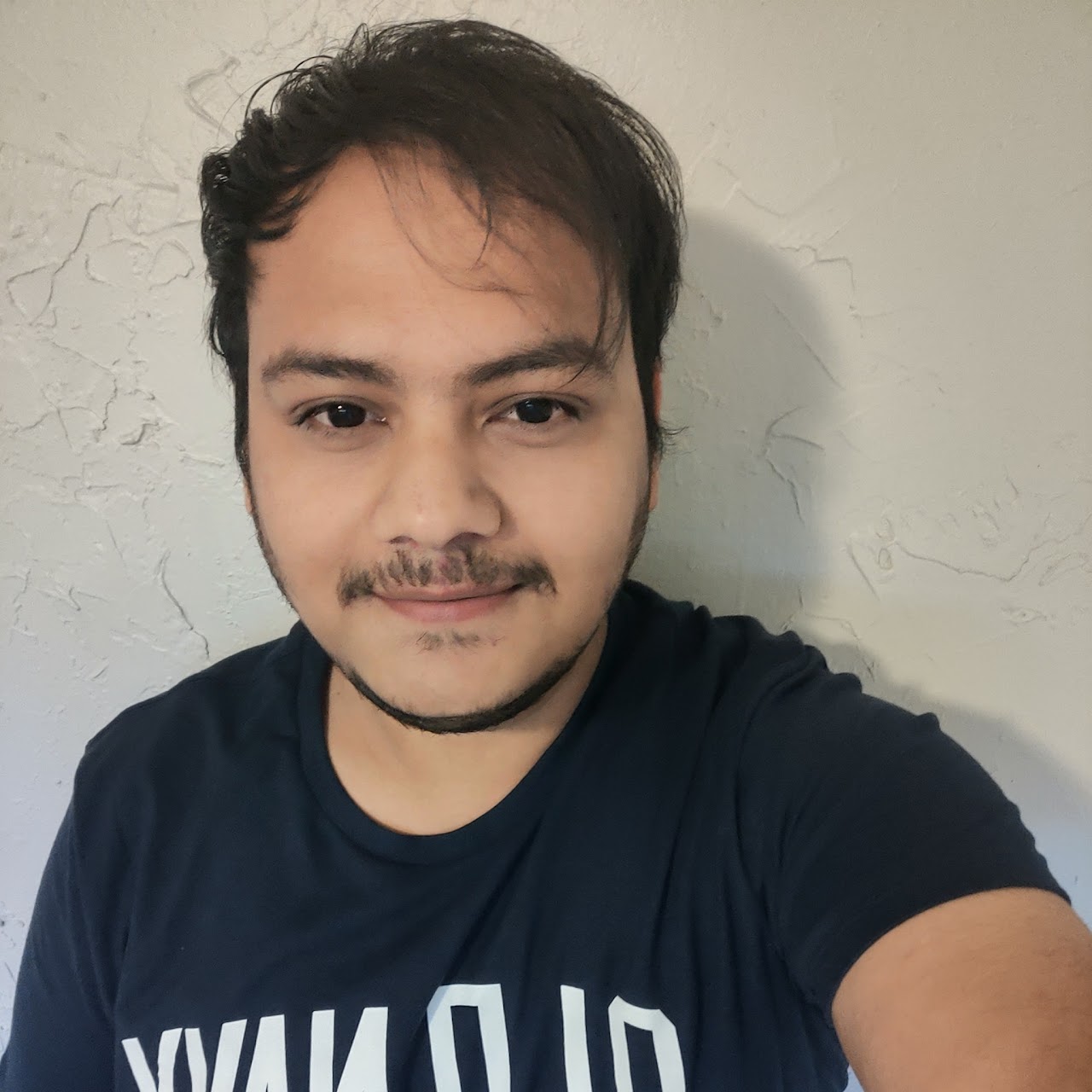}}]{Paras Sheth} received his B.Tech degree in Computer Sience and Engineering from Institute of Engineering and Management. He is currently a third-year PhD student of Computer Science and Engineering at Arizona State University. His research interests include causal inference, data mining and causal ML. Contact him at psheth5@asu.edu.
\end{IEEEbiography}

\begin{IEEEbiography}[{\includegraphics[height=1in,clip,keepaspectratio]{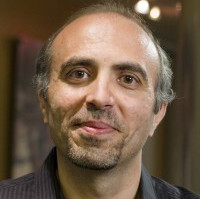}}]{K. Se\c{c}uk Candan}
received his Ph.D. degree in Computer Science from the University of Maryland College Park. He is currently a Professor of Computer
Science and Engineering and the Director of  Center for Assured and Scalable Data engineering (CASCADE) at Arizona State University. His research interests include data management, analysis, and scalable ML.  He is a Distinguished Scientist of the ACM.
\end{IEEEbiography}
\begin{IEEEbiography}[{\includegraphics[height=1.25in,clip,keepaspectratio]{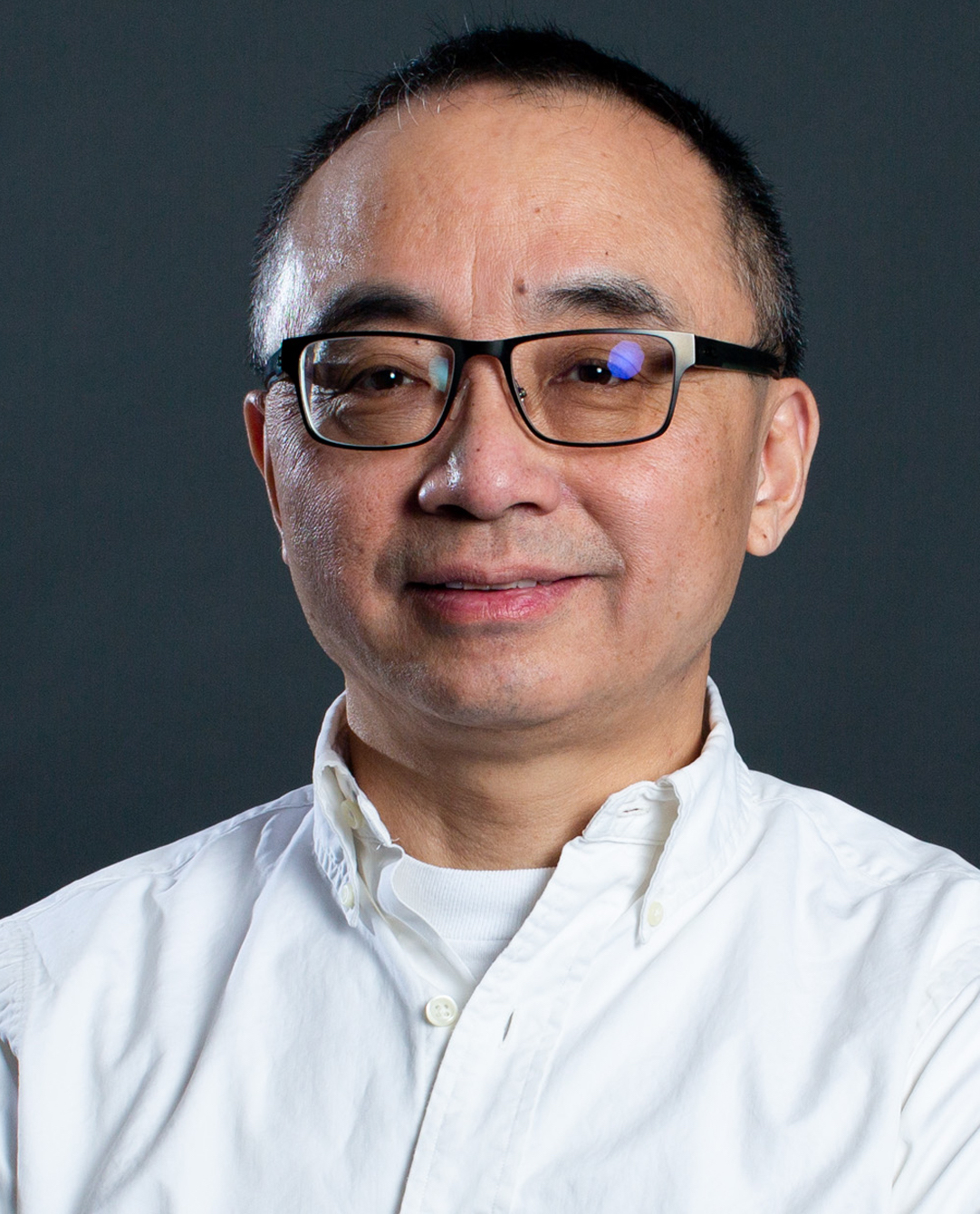}}]{Huan Liu}{\space}(F’12) received the B.Eng. degree in computer science and electrical engineering from Shanghai Jiaotong University and the Ph.D. degree in computer science from the University of Southern California. He is currently a Professor of computer
science and engineering at Arizona State University. His research interests include data mining, ML, social computing, and artificial intelligence, investigating problems that arise in
many real-world applications with high-dimensional data of disparate forms. His well-cited publications include books, book chapters, and encyclopedia entries and conference, and journal papers. He is a Fellow of IEEE, ACM, AAAI, and AAAS.
\end{IEEEbiography}

\end{document}